\def\year{2022}\relax
\documentclass[letterpaper]{article} % DO NOT CHANGE THIS
\usepackage{aaai22}  % DO NOT CHANGE THIS
\usepackage{times}  % DO NOT CHANGE THIS
\usepackage{helvet}  % DO NOT CHANGE THIS
\usepackage{courier}  % DO NOT CHANGE THIS
\usepackage[hyphens]{url}  % DO NOT CHANGE THIS
\usepackage{graphicx} % DO NOT CHANGE THIS
\usepackage{natbib}  % DO NOT CHANGE THIS AND DO NOT ADD ANY OPTIONS TO IT
\usepackage{caption} % DO NOT CHANGE THIS AND DO NOT ADD ANY OPTIONS TO IT
\DeclareCaptionStyle{ruled}{labelfont=normalfont,labelsep=colon,strut=off} % DO NOT CHANGE THIS
\usepackage{hyperref}
\hypersetup{
    citecolor=blue,
    colorlinks=true,
    linkcolor=red,
    filecolor=magenta,      
    urlcolor=blue,
linktocpage}
\usepackage{algorithm}
\usepackage{algorithmic}
\usepackage[utf8]{inputenc} % allow utf-8 input
\usepackage[T1]{fontenc}    % use 8-bit T1 fonts
\usepackage{hyperref}       % hyperlinks
\usepackage{url}            % simple URL typesetting
\usepackage{booktabs}       % professional-quality tables
\usepackage{microtype}      % microtypography
\usepackage{subfigure}
\usepackage{adjustbox} 
\usepackage{threeparttable} % table footnote
\usepackage{amsmath} % enabling various math envs
\usepackage{amsfonts} % blackboard math symbols
\usepackage{amssymb} % enabling ams symbols (e.g. \mathbb), internally loading amsfont.sty
\usepackage{amsthm}  % providing {theorem} env
\usepackage{xfrac}
\newtheorem{theorem}{Theorem}
\newtheorem{lemma}{Lemma}

\newtheorem{assumption}{Assumption}
\newtheorem{proposition}{Proposition}
\newcommand\diff{\mathop{}\!\mathrm{d}}

\DeclareRobustCommand{\frac}[3][0pt]{{\begingroup\hspace{#1}#2\hspace{#1}\endgroup\over\hspace{#1}#3\hspace{#1}}} % fancier \frac, e.g. \frac[1.0pt]{1}{2}
\usepackage{bm}
\usepackage[noamssymbols,upint]{newtxmath} % various optional arguments
\usepackage[cal=euler,calscaled=1.0,scr=boondoxo,scrscaled=1.0]{mathalfa}
\usepackage{makecell} % enabling fancy commands for table cells
\usepackage{threeparttable} % table footnote
\usepackage{xcolor} % enabling more colors for comments
\makeatletter
\renewcommand*\env@matrix[1][c]{\hskip -\arraycolsep
  \let\@ifnextchar\new@ifnextchar
  \array{*\c@MaxMatrixCols #1}}
\makeatother
\usepackage{newfloat}
\usepackage{listings}
\floatstyle{ruled}
\newfloat{listing}{tb}{lst}{}
\floatname{listing}{Listing}
\title{Revisiting the Characteristics of Stochastic Gradient Noise and Dynamics}
\author{
    Yixin Wu$^1$\equalcontrib,
    Rui Luo$^1$\equalcontrib,
    Chen Zhang$^1$,
    Jun Wang$^1$,   
    Yaodong Yang$^2$
}
\affiliations{
    $^1$University College London, $^2$King's College London\\
    \{yixin.wu, rui.luo, chen.zhang, jun.wang\}@cs.ucl.ac.uk, yaodong.yang@kcl.ac.uk
}

\usepackage{bibentry}
\begin{document}

%%%%%%%%%%%%%%%%%%%%%%%%%%%%%%%%%%%%%%%%%%%%%%%%%
\maketitle
\begin{abstract}

In this paper, we characterize the noise of stochastic gradients and analyze the noise-induced dynamics during training deep neural networks by gradient-based optimizers. Specifically, we firstly show that the stochastic gradient noise possesses finite variance, and therefore the classical Central Limit Theorem (CLT) applies; this indicates that the gradient noise is asymptotically Gaussian. Such an asymptotic result validates the wide-accepted assumption of Gaussian noise. We clarify that the recently observed phenomenon of heavy tails within gradient noise may not be intrinsic properties, but the consequence of insufficient mini-batch size; the gradient noise, which is a sum of limited \emph{i.i.d.} random variables, has not reached the asymptotic regime of CLT, thus deviates from Gaussian. We quantitatively measure the goodness of Gaussian approximation of the noise, which supports our conclusion. Secondly, we analyze the noise-induced dynamics of stochastic gradient descent using the Langevin equation, granting for momentum hyperparameter in the optimizer with a physical interpretation. We then proceed to demonstrate the existence of the steady-state distribution of stochastic gradient descent and approximate the distribution at a small learning rate. 
% Finally, we revisit the characteristics of dynamics for escaping local minima, under the framework of Kramers' escape rate with inhomogeneous medium and anisotropic noise.

% In this paper, we analyze the characteristics of the noise within the stochastic gradient during the training procedure of deep neural networks using gradient-based optimizers. Firstly, we proved that the gradient noise possesses finite variance and therefore the classical Central Limit Theorem applies, which indicates that the noise is asymptotically Gaussian. This leads to the conclusion that the observations on the gradient noise exhibiting heavy-tailed phenomenon is due to the insufficient size of mini-batches: the noise as a sum of limited \emph{i.i.d.} random variables has not reached the asymptotic regime. Secondly, we evaluate in a quantitative sense the goodness of Gaussian approximation of the gradient noise and then establish a relation between the Gaussianity and the standardized third absolute moment in the Berry-Esseen theorem.
\end{abstract}

\section{Introduction}

A typical machine learning task involves the minimization of an empirical loss function evaluated over instances within a specific training dataset. When the representation model is built on deep neural networks, millions of parameters are to be optimized; it is a common practice that one invokes a gradient-based optimization method to solve for the minimum of interest \cite{lecun2015deep}. On the other hand, the datasets for training have been expanding to a very large scale, where the algorithms equipped with stochastic gradients are de facto dominating the practical use of gradient-based optimizers. Therefore, stochastic gradient optimizers have been of significant interest in the community of machine learning research in the era of deep learning, which facilitates various applications of deep learning to real-world problems.

Originated from the stochastic approximation \cite{robbins1951stochastic,kushner2003stochastic}, various algorithms have been proposed and gained credits for the triumph of deep learning in a broad range of applications, including but not limited to finance \cite{kim2020can,zhang2018benchmarking}, gaming AI \cite{silver2016mastering,peng2017multiagent,yang2020overview}, autonomous driving \cite{grigorescu2020survey,zhou2020smarts} and protein structure discovery \cite{jumper2021highly}. The momentum methods were conceived along with the advent of back-propagation \cite{rumelhart1986learning}, which has been proved to have a fundamental physical interpretation \cite{qian1999momentum} and is very important for fast convergence \cite{sutskever2013importance}. Adaptive methods, e.g. AdaGrad \cite{duchi2011adaptive}, RMSProp \cite{tieleman2012lecture} and Adam \cite{kingma2014adam}, have demonstrated very fast convergence and robust performance, and now dominating the practical applications. Albeit greatly successful, it may still need more insights on the training dynamics guided by the gradient-based methods in a systematic perspective when one trains deep neural networks for real-world tasks.

% \rui{Adam traces running variance of the stochastic gradients, if infinite, then it fails}

Investigations devoted to this topic typically presume the stochastic gradients converge as discretizations of continuous-time dynamical systems, where the dynamical properties of the continuous-time counterpart are studied. The specification of the dynamical systems then lies in identifying the gradient noise. A classical assumption regarding the gradient noise is that it converges asymptotically towards Gaussian random variables due to the classical Central Limit Theorem (CLT)~\cite{feller:1971,chung2001course}, behind which the variance of stochastic gradients are thought to be finite, and the size of mini-batch is reasonable enough for a good Gaussian approximation. Such assumption applies to typical variational analysis regarding SGD and its variants~\cite{mandt2017stochastic}. Based on Gaussian noise assumption, some works approximate the dynamic of SGD by Langevin dynamics and proved its escaping efficiency from local minima~\cite{li2017stochastic,he2019control,zhou2019toward,hu2019diffusion,luo2018thermostat,luo2020replica}. These works assume that the noise covariance of SGD is constant or upper bounded by some constant.

Recently, the structure of the gradient noise has attracted more attention from the community. Many works study SGD from the perspective of its anisotropic nature of noise~\cite{zhu2019anisotropic,wu2020noisy,xie2020diffusion}. Empirical evidence~\cite{simsekli2019tail} has also been discovered that the gradient variance may be infinite, and it is the infinite variance that violates the conditions of the classical CLT with the Gaussian approximation. They introduced the generalized CLT which showed that the law of the sum of these i.i.d. variables with infinite variance still converges, but to a different family of heavy-tailed distributions.

We attribute such heavy tails phenomenon as the result of insufficient batch size for the noise (\emph{i.e.} a sum of limited \emph{i.i.d.} random variables) to converge to Gaussian rather than the hypothesis that the noise is heavy-tailed by its nature \cite{simsekli2019tail,simsekli2020fractional}. 
The investigation is taken through a different lens by evaluating the Gaussianity of noise in terms of the percentage of Gaussian elements for learnable weights in layers during the training procedure with different batch sizes: the gradient noise still exhibits good Gaussianity when the batch size increases. That is, the larger batch size we train, the higher Gaussianity for gradient noise we observe. Notably, there are some similar results showing in~\cite{panigrahi2019non} where they observe Gaussian behavior for gradient noise distribution at large batch size (over 500). To gain a better understanding and a more quantitative measurement, we involve the Berry-Esseen theorem which characterizes the rate of convergence for CLT~\cite{chung2001course} and then establish a link between the rate of convergence and the magnitude of the square root of kurtosis of the noise. With such evidence of Gaussian noise, we analyze the dynamics of SGD and unify the discretization scheme w.r.t. the general Langevin equations: the vanilla SGD corresponds to the high-friction limit of the continuous-time dynamics. The learning rate essentially corresponds to the effective temperature of the associated physical system. We provide a simple local approximation of the distribution of learning parameters when the dynamics enter the vicinity of minima.

% We emphasize that the intention of such investigation is not to validate the use of larger batch sizes during training in practice, we evaluate the Gaussianity on large batch sizes in order to provide evidences to support our claim that the gradient noise is not heavy-tailed by its nature but caused by the insufficient batch size that the noise is not converged. 

% These works studies the generalization influence of the magnitude of the SGD noise, which is controlled by the quotient of learning rate and batch size~\cite{hoffer2017train,jastrzkebski2017three}.

The main contribution of this paper is twofold.
1. We demonstrate that the noise within stochastic gradients possesses finite variance during the training procedure of deep neural networks, and therefore the noise is asymptotically Gaussian as is generally conceived due to the application of the classical Central Limit Theorem. We established a link of the rate of convergence to the kurtosis of the noise which can be readily computed.
2. We analyze the dynamics of stochastic gradient descent and the distribution of parameters, clarifying that the distribution deviates from the Gibbs distribution in the traditional settings. We provide a local approximation to the distribution when the parameters are close to optimal, which resembles an isotropic quadratic potential.

\section{Analysis of Gaussian Behavior}

\label{analysis}
In this section, we show that the gradient noise would not admit infinite variance during neural network training with the analysis on forward and backward propagation under reasonable assumptions.
And we further explain the non-Gaussianity phenomenon from the perspective of the Berry-Esseen theorem.

% \subsection{Notation}

% \begin{itemize}
% \item Distribution of the data $\mathscr{D}(x)$

% \item Randomly sampled datum $x_j \sim \mathscr{D}$

% \item Mini-batch $\mathscr{S} = \{x_j\}$

% \item $x_j$ and $x_{j'}$ are \emph{i.i.d.} random variables (\emph{r.v.}'s)

% \item Loss function per datum $\ell(x;\theta)$

% \item Empirical loss $\mathscr{L} \triangleq \sum_{x \in \mathscr{S}} \ell(x;\theta) / |\mathscr{S}|$

% \item activation function $v = f(u)$

% \item affine transformation $u = Wv + b$

% \item absorb $b$ into $W$

% \item Forward propagation $u^{(k)} = W^{(k)}v^{(k-1)} + b^{(k)}$

% \item $v^{(k)} = f_k\big(u^{(k)}\big)$ $f_k = f^{(k)}\big(u^{(k)}\big)$ and $f'_k = \diff\big[f^{(k)}(u^{(k)})\big]$

% \end{itemize}

\subsection{Preliminaries}
The optimization problems for neural networks usually involve minimizing an empirical loss with training data $\left\{x_{s}, y_s\right\}_{s=1}^{S}$ over the network parameter $\theta \in \mathbb{R}^{d}$.
To be more specific, we look for an optimal $\theta^\star$, st.
\begin{align}
    \theta^{\star}=\underset{\theta \in \mathbb{R}^{d}}{\arg \min }\left\{\mathscr{L}(\theta) \triangleq \frac{1}{|\mathscr{S}|} \sum_{x \in \mathscr{S}} \ell(x;\theta)\right\},
\end{align}

where $\mathscr{L}: \mathbb{R}^{d} \rightarrow \mathbb{R}$ denotes the empirical loss function in $\theta$, each $\ell(x;\theta)$ denotes the loss over one sample. 
% Note that for unsupervised learning problems, $y_i$ in above discussions will be omitted.
Stochastic gradient descent (SGD) is one of the most popular approaches for attacking this problem in practice. 
% It first randomly draws a mini-batch of samples with index set $B_t = \{i_1,...,i_b\}$, 
It performs parameter update $\theta_{t+1}=\theta_{t}-\alpha \nabla\mathscr{L}(\theta_t)$, using the stochastic gradient $\nabla\mathscr{L}(\theta)$ estimated from the mini-batch $\mathscr{S} = \{x_s\}$ and the learning rate $\alpha$, where
\begin{align}
\nabla\mathscr{L}(\theta) = \frac[1pt]{1}{|\mathscr{S}|} \sum_{x \in \mathscr{S}}\frac{\partial\ell(x;\theta)}{\partial{\theta}}.
\end{align}

The stochastic gradient noise, due to the fact that the gradients can be evaluated on different mini-batches, can be interpreted from additive and multiplicative viewpoints~\cite{wu2019multiplicative}. 
In this paper, we adopt the additive formulation and use $\epsilon$ to represent the additive noise, 
\begin{align}
\epsilon \triangleq \nabla\mathscr{L}(\theta) - \nabla\mathbb{E}\ell(x;\theta).
\end{align}

While $\theta$ assembles all the parameters in the feed forward neural network, we use $W^{(k)}$ and $b^{(k)}$ as weight matrix and bias vector in the $k$-th layer. Feed forward procedure is
$u^{(k)} = W^{(k)}a^{(k-1)} + b^{(k)}$, where $a^{(k)}=f_k\big(u^{(k)}\big)$ denotes the input of the $k$-th layer (post-activation of the ($k$-1)-th layer), $u^{(k)}$ denotes the pre-activation of the $k$-th layer and $f_k$ denotes the activation function entrywisely applied on the pre-activation. 
The updating rule for an entry $W_{ij}^{(k)}$ (at the i-th row and the j-th column) in the weight matrix $W^{(k)}$ is 
\begin{align}
 W_{ij}^{(k)} = W_{ij}^{(k)} - \alpha \frac{\partial\mathscr{L}}{\partial W_{ij}^{(k)}}.
\end{align}

\subsection{Central Limit Theorem}
The distinction between the requirements of CLT and Generalized CLT is whether the variances of summed variables are finite.
In this section, we analyze the variance of gradient noise during neural network training and thus show that the CLT still holds under mild assumptions. Firstly we provide mathematical induction to show that the variance of pre-/post-activations will not be infinite through the forward propagation. Next, we analytically show that the partial derivative of loss function w.r.t. the output of the neural network and the derivatives of loss function w.r.t. weights will not admit infinite variance.

Note that different steps in the proof have different minimum requirements on the data distribution for the conclusions to hold true.
The strongest assumption on the data in our discussion is that the data distribution has finite fourth order moments.
While this is automatically true for empirical distributions, which have been adopted in the literature ~\cite{jacot2018neural}, it is also true for discrete or continuous distributions with bounded support, e.g., image data with range $[0, 255]$. Preceding the full proof, we show the following lemma for Lipschitz functions which will be used for the variance analysis of activation functions.\\

\noindent\textbf{Zero mean for SGD noise}\\
Deterministic, full-batch gradient:
\begin{align}
\phi(\theta) = \nabla\mathbb{E}\ell(x;\theta) = \int_x \frac{\partial\ell(x;\theta)}{\partial{\theta}} \diff{\mathscr{D}(x)}.
\end{align}
Stochastic, mini-batch gradient:
\begin{align}
\nabla\mathscr{L}(\theta) = \frac[1pt]{1}{|\mathscr{S}|} \sum_{x \in \mathscr{S}}\frac{\partial\ell(x;\theta)}{\partial{\theta}} \approx \int_x \frac{\partial\ell(x;\theta)}{\partial{\theta}} \diff{\mathscr{D}(x)},
\end{align}
\begin{align}
\mathbb{E}[\nabla\mathscr{L}(\theta)] &=
\mathbb{E}\bigg[\frac[1pt]{1}{|\mathscr{S}|} \sum_{x \in \mathscr{S}} \frac{\partial\ell(x;\theta)}{\partial{\theta}}\bigg] \nonumber\\
&=\frac[1pt]{1}{|\mathscr{S}|} \sum_{x \in \mathscr{S}} \nabla\mathbb{E}\ell(x;\theta) = \phi(\theta).
\end{align}

So we can obtain the expectation of stochastic gradient noise
$\mathbb{E}[\epsilon] = \mathbb{E}[\nabla\mathscr{L} - \mathbb{E}[\nabla\mathscr{L}(\theta)]] = \mathbb{E}[\nabla\mathscr{L} - \phi] = 0$.\\

\noindent\textbf{Finite variance for SGD noise} 
\begin{lemma}\label{lm:lip}
Assume that a random vector $x\in\mathbb{R}^d$ has finite variance on each entry. $f:\mathbb{R}\rightarrow\mathbb{R}$ is a Lipschitz function with some constant $c>0$ and entry-wisely applied on $x$.
Then the random vector $f(x)\in\mathbb{R}^d$ has finite variance on each entry.
\end{lemma}
The feed forward process could be regarded as recurrent applications of activations and linear transformations.
With the finite fourth moments assumption, we know the second order moments of the input data $a^{(0)} :=x$ are finite.
If the input to the $k$-th layer, i.e., $a^{(k-1)}$, has finite variance, we show that $a^{(k)}$ has finite variance.
Denote the mean and co-variance of $a^{(k-1)}$ respectively by $\bar{a}^{(k-1)}$ and $\Sigma^{(k-1)}$.
Recall $u^{(k)}=W^{(k)}a^{(k-1)}+b^{(k)}$, the mean and covariance of $u^{(k)}$ would be
\begin{align}
\mathbb{E}[u^{(k)}] &= W^{(k)}\bar{a}^{(k-1)}+b^{(k)},\\ \mathbf{cov}[u^{(k)}] &= W^{(k)}\Sigma (W^{(k)})^\top
\end{align}
Clearly, the variance of $u^{(k)}$, as part of the covariance $\mathbf{cov}[u^{(k)}]$, is finite.
We apply non-linearity $f_k(\cdot)$ to $u^{(k)}$ element-wisely and get the activation $a^{(k)}$, i.e. $a^{(k)}=f_k(u^{(k)})$.
For commonly adopted activations, i.e., Sigmoid, ReLU, LeakyRelu, Tanh, SoftPlus, etc, they are all Lipschitz functions.
With Lemma \ref{lm:lip}, the variance of $a^{(k)}$ is also finite.
Applying this logic recursively, we would know for the whole neural network, variances of all pre-/post-activations are finite. The variance of the final output $a^{(K)}$ is finite as well.

To use induction for back propagation, we show the variance of the derivative of loss function w.r.t. $a^{(K)}$ is finite.
Here we investigate two variants of loss functions, i.e., mean square error loss function for regression problems and cross-entropy loss function for classification problems.
The mean square error loss is defined as $\mathscr{L}_{MSE} = ||y-a^{(K)}||^2$.
Since $\partial \mathscr{L}_{MSE}/\partial a^{(k)}=-2(y-a^{(k)})$, it is obvious that it has finite variance with our assumption on data and above induction.
To complete the discussion, we show the results for cross-entropy loss in the following lemma.
\begin{lemma}
For cross entropy loss function $\mathscr{L}_{CE} = -\sum_{j=1}^M y_j\log\big[\exp{a^{(K)}_j}/\sum_{k}\exp{a^{(K)}_k}\big]$, the gradient w.r.t. the network output $a^{(K)}$, i.e., $\partial \mathscr{L}_{CE}/\partial a^{(K)}$, as a function on random variables $a^{(K)}$ and label $y$, has finite variance.
\end{lemma}
The derivatives of the loss function w.r.t. each entry of the weight matrix admit the following forms by the chain rule,
\begin{align}\label{eq:grad_w}
\frac[1pt]{\partial\mathscr{L}}{\partial{W^{(k)}_{ij}}} = \frac[1pt]{\partial\mathscr{L}}{\partial{u^{(k)}_{i}}}\frac[1pt]{\partial{u^{(k)}_{i}}}{\partial{W^{(k)}_{ij}}}.
\end{align}
Note that the partial gradient $\partial\mathscr{L}/\partial W_{ij}^{(k)}$ consists of both global information, i.e., $\partial\mathscr{L}/\partial u^{(k)}_i$, and local information, i.e.,  $\partial u^{(k)}_i/\partial W_{ij}^{(k)}$. For the global information, there holds the following recurrent relationship
\begin{align}\label{eq:grad_recurrent}
\frac[1pt]{\partial\mathscr{L}}{\partial{u^{(k-1)}}} = \bigg( \big[W^{(k)}\big]^\top \frac[1pt]{\partial\mathscr{L}}{\partial{u^{(k)}}} \bigg) \odot f'_{k-1}\big(u^{(k-1)}\big),
\end{align}
% \begin{equation}\label{eq:grad_recurrent}
%     \frac{\partial\mathscr{L}}{\partial u^{(k-1)}}=f^\prime_{k-1}(u^{(k-1)})\odot\left((W^{(k)})^T\frac{\partial\mathscr{L}}{\partial u^{(k)}}\right),
% \end{equation}
where $\odot$ denotes the entrywise Hadamard product. Since $\partial\mathscr{L}/\partial a^{(K)}$ has finite variance and Lipschitz continuous activation functions have bounded (sub-)gradients, $\partial\mathscr{L}/\partial u^{(K)}$ also has finite variance.
From Equation~\ref{eq:grad_recurrent} we know $\partial\mathscr{L}/\partial u^{(k-1)}$ has finite variance.
By mathematical induction we conclude that $\partial\mathscr{L}/\partial u^{(k)}$ has finite variance for all $k$. For the local information, 
\begin{align}
\label{eq:u_tau}
\frac[1pt]{\partial u_{\tau}^{(k)}}{\partial W^{(k)}_{ij}}=
\begin{cases}
\,a_{j}^{(k-1)}& \text{$\tau      =      i$}\\
\,0& \text{$\tau      \neq      i$}
\end{cases}
\end{align}
$\partial u^{(k)}/\partial W_{ij}^{(k)}$ is a vector with the $i$-th entry being $a_j^{(k-1)}$ and other entries being $0$'s. 
% while $\sfrac{\partial u^{(k)}}{\partial b_i^{(k)}}$ is a vector with the $i$-th entry being $1$ and other entries being $0$'s.
% Since $\sfrac{\partial u^{(k)}}{\partial b_i^{(k)}}$ only contains one non-zero entry which is $1$, its variance essentially comes from the variance of the global information $\sfrac{\partial\mathscr{L}}{\partial u^{(k)}}$, which is finite.
% In $\sfrac{\partial\mathscr{L}}{\partial b_i^{(k)}}$, $\sfrac{\partial u^{(k)}}{\partial b_i^{(k)}}$ also only contains one non-zero entry $v^{(k-1)}_j$, but this non-zero entry is a random variable.
With the finiteness assumption on fourth order moments, we can employ Cauchy–Schwarz inequality to bound the variance of $\partial\mathscr{L}/\partial W^{(k)}_{ij}$, c.f. the supplemental material. 
Since gradient noise is the centered version of the partial gradient $\partial\mathscr{L}/\partial W^{(k)}_{ij}$
% or $\sfrac{\partial\mathscr{L}}{\partial b^{(k)}_i}$
, it will also have finite variance.

\subsection{CLT convergence rate}

In the previous subsection, we showed that the variance of gradient noise is finite under reasonable assumptions on the data and support that CLT will hold.
% Take $\partial\mathscr{L}/\partial W^{(k)}_{ij}$ as an example.
% The gradient evaluated on a mini-batch with indices $S=\{i_1,\ldots,i_s\}$ is the average of $b$ i.i.d. random variables, i.e., 
% \begin{equation}
%     \frac{1}{b}\bigg[\frac{\partial\mathscr{L}}{\partial W^{(k)}_{ij}}(x_{i_1}, y_{i_1}) + \ldots + \frac{\partial\mathscr{L}}{\partial W^{(k)}_{ij}}(x_{i_s}, y_{i_s})\bigg].
% \end{equation}
% Subtracting the mean on the whole data distribution from the above average, 
% the stochastic gradient noise would have zero mean, finite variance and finite third absolute moment under the same assumptions on data.
Then the theorem about convergence rate of CLT, i.e. Berry–Esseen theorem \cite{feller:1971,chung2001course} applies.
\begin{theorem}[Berry–Esseen theorem]
There exists a universal constant $A_0$, s.t. for any i.i.d. random variables $\{\chi_s\}_{s=1}^S$ with $\mathbb{E}[\chi_s]=0$, $\mathbf{var}[\chi_s]=\sigma^2$ and $\beta := \mathbb{E}[|\chi_s|^3]<\infty$, the following inequality holds
\begin{align}\label{eq:betheorem}
    \sup_\chi |F_{|\mathscr{S}|}(\chi)-\Phi(\chi)| \le \frac{\beta}{\sigma^3}\frac{A_0}{\sqrt{|\mathscr{S}|}},
\end{align}
where $\Phi$ is the c.d.f. of standard Gaussian distribution and $F_{|\mathscr{S}|}$ is the c.d.f. of random variable $\sum_{s=1}^S\chi_s/\sqrt{|\mathscr{S}|}\sigma$.
\end{theorem}

More specifically, assume that neural network is a deterministic function, $x$ is an input datum, $\ell(x;\theta)$ is a loss function per datum, $u^{(k)}(x)$ is a pre-activation at $k$-th layer. By Equation~\ref{eq:grad_recurrent}, we can have
\begin{align}
\frac[1pt]{\partial\mathscr{L}}{\partial{u^{(k)}}}(x) = \bigg( \big[W_{ij}^{(k+1)}\big]^\top \frac[1pt]{\partial\mathscr{L}}{\partial{u^{(k+1)}}}(x) \bigg) \odot f'_{k}\big(u^{(k)}(x)\big),
\end{align}
which is a function of $x$. The mini-batch gradient ($x_s \in \mathscr{S}$)
\begin{align}
\nabla\mathscr{L}(\theta) = \frac[1pt]{1}{|\mathscr{S}|} \sum_{s}g_s = \frac[1pt]{1}{|\mathscr{S}|} \sum_{s}\frac[1pt]{\partial\ell(x_s)}{\partial{W_{ij}^{(k)}}}.
\end{align}
Recall Equation~\eqref{eq:grad_w} \& \eqref{eq:u_tau}, we have 
\begin{align}
g_s = \frac[1pt]{\partial\ell(x_s)}{\partial{W_{ij}^{(k)}}} = \sum_{\tau}\frac[1pt]{\partial\ell(x_s)}{\partial{u^{(k)}_{\tau}}}\frac[1pt]{\partial{u^{(k)}_{\tau}}}{\partial{W^{(k)}_{ij}}} = \frac[1pt]{\partial\ell(x_s)}{\partial{u^{(k)}_{i}}}a_{j}^{(k-1)}
\end{align}
For each data point, the gradient noise $\triangle g = g - \mathbb{E}[g]$ is zero mean. From Inequality Equation~\eqref{eq:betheorem}, we know that the convergence of CLT is bounded by three quantities. Since $A_0$ is the universal constant and $|\mathscr{S}|$ is the batch size, we need to estimate the standardized third absolute moment $\beta/\sigma^3$. We define
\begin{align}
\beta &= \mathbb{E}[|\triangle g|^3],\\
\sigma^2= \mathbf{var}[\triangle g] &= \mathbf{var}[g] = \mathbf{var}\bigg[\frac[1pt]{\partial\ell(x_s)}{\partial{u^{(k)}_{i}}}a_{j}^{(k-1)}\bigg].
\end{align}
Since the third absolute moment is hard to compute, with $\sigma^2= \mathbb{E}[(\triangle g)^2]$, we provide a ready upper bound using Cauchy-Schwarz Inequality,
\begin{align}
\frac[1pt]{\beta}{\sigma^3} &= 
\frac[1pt]{\mathbb{E}[|\triangle g(\triangle g)^2|]}{\sigma^3}\leq
\frac[1pt]{\sqrt{\mathbb{E}[(\triangle g)^2]}\sqrt{\mathbb{E}[(\triangle g)^4]}}{\sigma^3}\nonumber\\&=
\frac[1pt]{\sigma \sqrt{\mathbb{E}[(\triangle g)^4]}}{\sigma^3}=
% \frac[1pt]{\sqrt{\mathbb{E}[(\triangle g)^4]}}{\sigma^2} \\
% &= \sqrt{\frac[1pt]{\mathbb{E}[(\triangle g)^4]}{\sigma^4}}
\sqrt{\mathbb{E}\bigg[\bigg(\frac[1pt]{\triangle g}{\sigma}\bigg)^4\bigg]}
\end{align}
We denote standard score $z = \triangle g/ \sigma$ which is a random variable with zero mean and unit variance. By the interpretation of kurtosis from~\cite{moors1986meaning},
\begin{align}
\mathbb{E}\left[z^{4}\right]&=\mathbf{var}\left[z^{2}\right]+\left[\mathbb{E}\left[z^{2}\right]\right]^{2}\nonumber\\
&=\mathbf{var}\left[z^{2}\right]+[\mathbf{var}[z]]^{2}=\mathbf{var}\left[z^{2}\right]+1
\end{align}
Finally we have the upper bound
\begin{align}
\frac[1pt]{\beta}{\sigma^3} \leq  \sqrt{\mathbf{var}\left[z^{2}\right]+1}
\end{align}
% \begin{align}
% \sigma^3 &= \big[\sqrt{\mathbf{Var}[\triangle g]}\big]^3 = \big[\sqrt{\mathbf{Var}[g]}\big]^3
% \end{align}

% \begin{align}
% \sigma^2= \mathbf{var}[\triangle g] = \mathbf{var}[g] = \mathbf{var}\bigg[\frac[1pt]{\partial\ell(x_s)}{\partial{u^{(k)}_{i}}}a_{j}^{(k-1)}\bigg]
% \end{align}
% where $W$ is a matrix with size $M\times N$, $a$ is a vector with $N \times 1$ and $u$ is a vector with $M \times 1$. This variance value can be estimated by variance of products method~\cite{goodman1960exact,bohrnstedt1969exact}

% \begin{align}
% \mathbb{E}[(\triangle g)^4] &=\mathbf{Var}[(\triangle g)^2] + \big(\mathbb{E}[(\triangle g)^2]\big)^2 \\&=
% \mathbf{Var}[(\triangle g)^2] + \sigma^4
% \end{align}
Standing on this ground, we assert that the non-Gaussianity of gradient noise observed in some neural network training is due to the fact that the asymptotic regime has not been reached with that settings.
In the following section, we also support our statements with some numerical experiments.

\section{Dynamics of SGD}

Given gradient noise is asymptotically Gaussian, the dynamics of gradient descent resembles the motion of a particle within a potential field. The topic can be related to the classical Kramers Escape Problem~\cite{kramers1940brownian}.
% [Rumelhart 1986] Eq. 9
% \begin{align}
% \Delta{w(t)} = -\varepsilon \partial{E} / \partial{w}(t) + \alpha w(t-1)
% \end{align}
% dynamics of gradient descent [Sutskever 2013]
% \begin{align}
% v_t &= \mu v_{t-1} - \varepsilon \nabla f(\theta_{t-1} + \mu v_{t-1})\notag\\
% \theta_t &= \theta_{t-1} + v_t
% \end{align}
% \begin{align}
% \Delta{\theta} = v\Delta{t}
% \end{align}
% \begin{align}
% \diff{\theta} &= v\diff{t}\notag\\
% m\diff{v} &= - \gamma v - \nabla V(\theta) - \sqrt{2\beta kT}\epsilon(t) \diff{t}
% \end{align}
Consider the procedure of gradient descent optimizers, the parameter vector $\theta \in \mathbb{R}^d$ is updated according to the gradient of loss, which resembles the evolution of position of a particle within the external position-dependent potential $\phi(\theta) = \mathbb{E}\ell(\theta)$. With gradient being noisy, the latter is typically described by the Langevin equation and the stochastic gradient noise is effectively the random kicks acting on the particle
\begin{align}
\label{eq:langevin}
m\frac[1pt]{\diff^2{\theta}}{\diff{t}^2} + \gamma\frac[1pt]{\diff{\theta}}{\diff{t}} = -\nabla\phi(\theta) + \eta(\theta, t)
\end{align}
where $\mathbb{E}\ell(\theta)$ is the average of gradients evaluated within a mini-batch also expressed as the underlying potential $\phi$, and $\eta(\theta, t) \in \mathbb{R}^d$ is $\theta$-dependent $\delta$-correlated random fluctuation. $\mathbb{E}[\eta(\theta, t)\eta(\theta, t')^\top] = \mathsf{D}(\theta)\delta(t - t')$ with $\mathsf{D}(\theta)$ denoting the corresponding $\theta$-dependent diffusion matrix. Discretization w.r.t. time translates the random kicks as
\begin{align}
\eta(\theta, t) \to \epsilon_n(\theta) \sim \mathscr{N}\big(0, \mathsf{C}(\theta)\big),
\end{align}
where we can express the covariance of SGN $\epsilon$ as $\mathsf{C}(\theta) = \mathsf{D}(\theta)/\Delta{t}$. The discretization of the Langevin equation \eqref{eq:langevin} w.r.t. time reads
{\small
\begin{align}
\label{eq:discret_langevin}
m\frac{\Delta^2\theta_{n}}{\Delta{t}^2} + \gamma\frac{\Delta\theta_{n}}{\Delta{t}} = -\nabla\mathbb{E}\ell(\theta) + \epsilon_{n-1}(\theta) = -\nabla\mathscr{L}(\theta_{n-1}).
\end{align}
}
Let us define $v_n = \Delta\theta_n = \theta_n - \theta_{n-1} = \theta(n\Delta{t}) - \theta((n-1)\Delta{t})$, substituting $v_n$ into Equation~\eqref{eq:discret_langevin} we can obtain
\begin{align}
m\frac{v_n - v_{n-1}}{\Delta{t}^2} + \gamma\frac[1pt]{v_n}{\Delta{t}} = -\nabla\mathscr{L}(\theta_{n-1})
\end{align}
By re-arranging terms, 
\begin{align}
v_n = \underset{\rho}{\underbrace{\frac{1}{1 + \gamma\Delta{t} / m}}} v_{n-1} - \underset{\alpha}{\underbrace{\frac{\Delta{t}}{\gamma + m/\Delta{t}}}} \nabla\mathscr{L}(\theta_{n-1}).
\end{align}
Defining momentum term as $\rho$ and learning rate as $\alpha$,
% \begin{align}
% \rho = \frac{1}{1 + \gamma\Delta{t} / m} \quad \mbox{and} \quad \alpha = \frac{\Delta{t}}{\gamma + m/\Delta{t}},
% \end{align}
we finally have
\begin{align}
\label{eq:sgd}
\begin{cases}
\, v_n = \rho v_{n-1} - \alpha \nabla\mathscr{L}(\theta_{n-1})\\
\, \theta_n = \theta_{n-1} + v_n
\end{cases}
\end{align}
which resembles the formulation of the classical momentum method~\cite{sutskever2013importance} or method of ``heavy-sphere''~\cite{polyak1964some}. For high-friction limit, i.e. $\gamma \gg m/\Delta{t}$, $\rho \to 0^+$, $\alpha \to \Delta{t} / \gamma$, Equation~\eqref{eq:sgd} can be simplified as the update rule for vanilla SGD
\begin{align}
\theta_n = \theta_{n-1} - \alpha \nabla\mathscr{L}(\theta_{n-1}).
% \mbox{~~~~with~~~~} \alpha = \Delta{t} / \gamma
\end{align}
This can be obtained alternatively by leveraging the \emph{overdamped} Langevin equation, where we omit the second derivative in Equation~\eqref{eq:langevin} as in high-friction limit the inertial force is negligible comparing to the friction.

% \begin{align}
% \partial_t{\pi} = - \varv^\top \nabla_{\theta}{\pi} + \nabla_{\varv}^\top \Big[\pi\:\!\big[\nabla\mathbb{E}\ell(\theta) + \gamma\varv\big]\Big] + \gamma \mathsf{D}(\theta) \odot \nabla_{\varv}^\top \nabla_{\varv}{\pi}
% \end{align}

Now we focus on the Langevin equation which is the analogue of momentum method in continuous time. Associated with the Langevin equation~\eqref{eq:langevin}, the equation describing the evolution of the distribution $\pi(\theta, \varv, t)$ is the well-known \emph{Kramers equation} with coordinates $\theta$ and velocity $\varv = \diff{\theta} / \diff{t} \in \mathbb{R}^d$. We write the equation in the form of \emph{continuity equation} \cite{risken1996fokker,van1992stochastic} as follows, 
\begin{align}
\label{eq:kramers}
\frac[1pt]{\partial{\pi}}{\partial{t}} -\,\mathbf{div}\,\mathcal{J} = 0
\end{align}
where we suppose the unit mass $m = 1$ and unity temperature, $\mathcal{J}$ is the probability current
\begin{align}
\label{eq:j}
\mathcal{J} =
\begin{Bmatrix}
\begin{bmatrix} 
-\varv\\[\jot]
\gamma\varv + \nabla\phi(\theta)
\end{bmatrix} + \frac[1pt]{1}{2}
\begin{bmatrix}
0\\
\mathsf{D}(\theta) \cdot \nabla_{\varv}
\end{bmatrix}
\end{Bmatrix}\pi(\theta, \varv, t)
\end{align}
However, one may not always find such a solution for the general Fokker-Planck equation~\cite{van1992stochastic}. For the steady-state distribution to exists, some conditions have to be imposed on the drift vector as well as the diffusion matrix; it can be shown that under mild conditions the existence of the steady-state distribution for \eqref{eq:kramers} can be found~\cite{soize1994fokker}.

\begin{proposition}
\label{p:detailed_balance}
The dynamics \eqref{eq:kramers} in general does not satisfy detailed balance. The steady-state distribution associated with \eqref{eq:kramers}, if exists, deviates from the classical Gibbs distribution in thermal equilibrium.
\end{proposition}

The deviation has been discovered recently in training neural networks~\cite{kunin2021rethinking}; Similar phenomena is observed in physics, which is often referred to as \emph{broken detailed balance}~\cite{battle2016broken,gnesotto2018broken}.

\begin{assumption}
\label{asm:c2}
The potential $\phi \in \mathcal{C}^2$ is second-order differentiable in the vicinity of mimimum $\theta_*$.
\end{assumption}
\begin{align}
\label{eq:hessian}
\mathsf{C}(\theta) = \mathbf{cov}[\nabla\mathscr{L}(\theta)] &= \nabla^2\phi(\theta) - \nabla\phi(\theta)\nabla^\top\phi(\theta) \notag\\
& \approx \nabla^2\phi(\theta_*) = \mathsf{H}(\theta_*) = \mathsf{H}_*
\end{align}
The first approximation is due to the fact that gradient noise variance dominates the gradient mean near critical points~\cite{zhu2019anisotropic,xie2020diffusion}. When evaluating at the true parameter $\theta_*$, there is the exact equivalence between the Hessian $\mathsf{H}$ and Fisher information matrix $\mathsf{F}=\nabla^2\phi(\theta_*)$ at $\theta_*$, referring to Chapter 8 of~\cite{pawitan2001all}.
% \begin{align}
% \mathsf{C}(\theta) &= \mathbf{cov}[\epsilon(\theta)] = \mathbf{cov}[\nabla\mathscr{L}(\theta)] \notag\\
% &= \mathbb{E}[\nabla^2\mathscr{L}(\theta)] - \mathbb{E}[\nabla\mathscr{L}(\theta)]\mathbb{E}[\nabla\mathscr{L}(\theta)]^\top \notag\\
% &= \nabla^2\mathbb{E}\ell(\theta) - \nabla\mathbb{E}\ell(\theta)\nabla^\top\mathbb{E}\ell(\theta) \notag\\
% &= \nabla^2\phi(\theta) - \nabla\phi(\theta)\nabla^\top\phi(\theta) \notag\\
% &= \nabla^2\phi(\theta) \notag\\
% & \approx \nabla^2\phi(\theta_*) = \mathsf{H}(\theta_*)
% \end{align}
According to~\cite{zhang2019algorithmic,xie2020diffusion}, we can approximate the potential in the vicinity of $\theta_*$ by its second-order Taylor series as
\begin{align}
\phi(\theta) \approx \phi(\theta_*) &+ \nabla^\top\phi(\theta_*) (\theta - \theta_*) \nonumber\\&+ \frac[1pt]{1}{2} (\theta - \theta_*)^\top \mathsf{H}_* (\theta - \theta_*)
\end{align}

Note that the diffusion matrix $\mathsf{D}(\theta) = \Delta{t}\mathsf{C}(\theta)$ is essentially small variables~\cite{xie2020diffusion,meng2020dynamic}. With $\Delta{t} \ll 1$, one may approximate the diffusion by a constant matrix with no dependency w.r.t. $\theta$ \cite{luo2017thermostat,luo2020replica}.

\begin{proposition}
\label{p:ss}
The steady-state distribution can be approximated when $\theta$ is in the vicinity of local minimum $\theta_*$, given Assumption \ref{asm:c2} applies (Proof in Appendix),
{\small
\begin{align}
\label{eq:steady-state}
\pi^{\mathrm{ss}} \propto \exp\Big[ -\frac[1pt]{\gamma}{\Delta{t}} (\theta - \theta_*)^\top(\theta - \theta_*) \Big] \exp\Big[ -\frac[1pt]{\gamma}{\Delta{t}}\varv^\top \mathsf{H}_*^{-1}\varv \Big].
\end{align}
}
\end{proposition}
In the scenario that we assume constant covariance in a small noise regime, the marginal distribution w.r.t. $\theta$ has the form of Boltzmann distribution with an efficient potential. If $\Delta{t} \ll 1$, Equation~\eqref{eq:steady-state} indicates that the probability peaks at the minima~\cite{hanggi1990reaction,matkowsky1977exit}. With the covariance as a variable, the distribution exists but may not have a closed form, but we believe the property of Proposition~\ref{p:ss} provides us with a good approximation.

% Due to the existence and uniqueness of solutions for the Fokker–Planck Equation, referring to Chapter 2 of~\cite{pavliotis2014fokker}, 

% The traces of SGN covariance matrices for the deep neural networks are very small non-constant values and can be well approximated by quadratic curves~\cite{meng2020dynamic}, the minimum of the quadratic curve is nearly located at the local minimum.

\section{Experimental Results}

In this experiment, we aim to verify the Gaussianity of stochastic gradient noise and find what kind of factors can affect the Gaussian behavior. We check the Gaussianity of stochastic gradient noise on different neural network architectures and conduct experiments on different models, namely AlexNet~\cite{krizhevsky2012imagenet} and Residual Neural Network (ResNet)~\cite{he2016deep} for classification problems. During the training, we test the Gaussianity of gradient noise for each layer. The data used for both networks is CIFAR-10\footnote{https://www.cs.toronto.edu/~kriz/cifar.html} dataset which consists of 60,000 colour images ($32 \times 32$ pixels) in 10 classes. There are 50,000 training images and 10,000 test images, with 6,000 images per class. 

In statistics, Gaussianity testing is a well-studied topic. The Shapiro and Wilk's $W$-test~\cite{shapiro1965analysis} is found to be the top test in a comparison study~\cite{razali2011power}. It assumes the null hypothesis that the given set of examples come from a Gaussian distribution. A small p-value provides strong evidence to reject the null hypothesis. The extension of the Shapiro-Wilk test for large samples up to $2000$ \cite{royston1982extension} (from the original sample size of 50). We use a revised algorithm called \textbf{AS R94} for fast calculation of the Shapiro-Wilk test \cite{royston1995remark} in our experiment. More specifically, in each layer we sample points from all the batches of stochastic gradient noise and test whether they are belonging to Gaussian distribution or not, then we calculate the percentage of Gaussianity for all weight dimensions to verify the Gaussianity assumption of gradient noise. During the experiment, we find some extreme cases where the gradient noise samples in one batch follow point mass behavior due to the ReLU activation function. We refer to these distributions as the Dirac delta function~\cite{dirac1981principles} which is the limit (in the sense of distributions) of the sequence of zero-centered Gaussian distributions. In the experimental results, the Shapiro-Wilk test also recognizes point mass data as Gaussian behavior.
It is worth noting that the image data have bounded support, and thus the gradient noise will not have infinite tails. As a result, it is not necessary to use tail index measuring to exam if the components rejected by the Gaussianity test have infinite variance or not -- the variances are finite by nature. See supplementary materials for more discussions related to the tail index.

\subsection{Gaussian behavior of AlexNet}
\label{Alexnet}

\begin{figure}[t]
\centering
\subfigure[Conv-layers in AlexNet]{
\label{fig:epoch_alexnet_1}
\includegraphics[width=0.475\columnwidth,height=3.3cm]{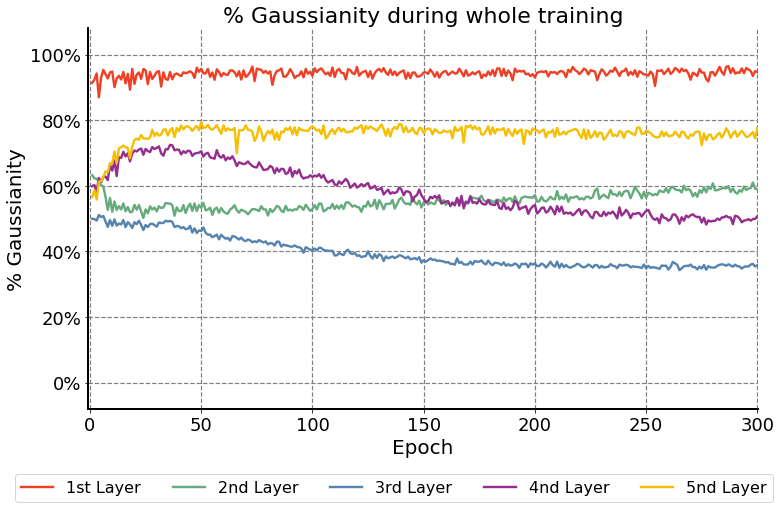}}
% \qquad
\subfigure[FC-layers in AlexNet]{
\label{fig:epoch_alexnet_2}
\includegraphics[width=0.475\columnwidth,height=3.3cm]{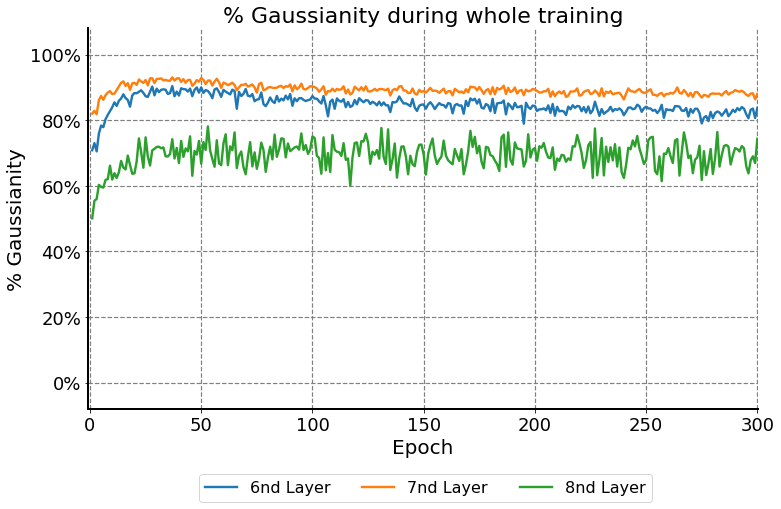}}
\vskip -5pt%

\subfigure[Conv-layers in AlexNet-BN]{
\label{fig:epoch_alexnet_bn_1}
\includegraphics[width=0.475\columnwidth,height=3.3cm]{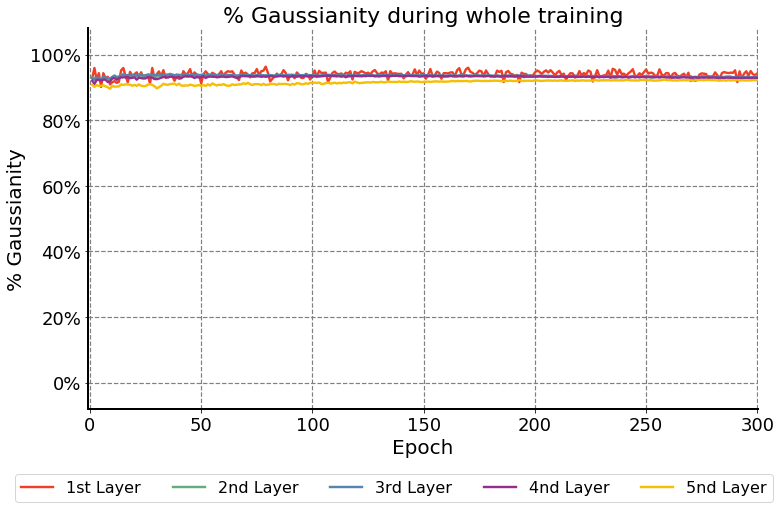}}
% \qquad
\subfigure[FC-layers in AlexNet-BN]{
\label{fig:epoch_alexnet_bn_2}
\includegraphics[width=0.475\columnwidth,height=3.3cm]{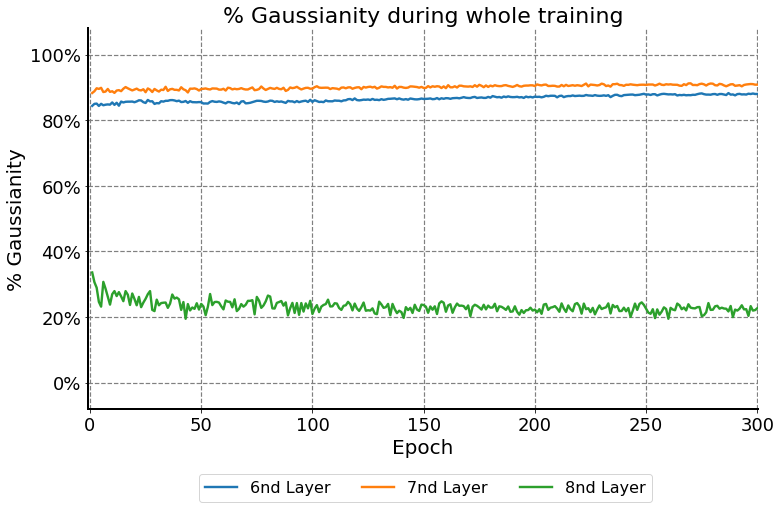}}
\vskip -5pt%
\caption{Gaussianity of each layer during training epochs.}
\vskip -10pt%
\label{fig:epoch_alexnet}
\end{figure}

AlexNet~\cite{krizhevsky2012imagenet} is one of the most successful convolutional neural architectures. In this experiment, we investigate the behavior of stochastic gradients during training. Instead of applying the original AlexNet tailored for ImageNet classification, modifications are necessary before applying AlexNet on datasets like Cifar-10 \cite{DBLP:conf/iclr/ZhangBHRV17}. We adopt some modifications to the architecture such that our network works with different formats of images. The modified AlexNet for Cifar-10 contains 8 learnable layers: the first 5 layers are 2$d$ convolutional layers (Conv-layers), which corresponds to the features extractor, the other 3 layers are fully-connected layers (FC-layers), comprising the classifier for prediction. We remove the \emph{local response} normalization in the feature extractor, the \emph{dropout} operations in the classifier, and the average-pooling layer between these two modules. We use non-overlapped max-pooling instead of the overlapped counterparts. The first convolutional layers are rescaled for adaptation on images of Cifar-10: kernel $5\times 5$, stride $2$, padding $2$. After this refinement of AlexNet for Cifar-10, we further composed a variant with batch normalization (BN), where we insert a batch-norm layer before each ReLU activation after the signal is transformed by either convolutional layers or fully-connected ones. 

In order to understand the change of Gaussian behavior of gradient noise during whole training epochs, we test classic AlexNet and AlexNet with batch-norm layers, show the Gaaussianity results of convolutional layers in Figure~\ref{fig:epoch_alexnet_1} \& \ref{fig:epoch_alexnet_bn_1} and results of fully connection layers in Figure~\ref{fig:epoch_alexnet_2} \& \ref{fig:epoch_alexnet_bn_2}. During total 300 epochs training, we set learning rate is 0.1 before 150 epochs, 0.01 within 150 epochs and 225 epochs, 0.001 from 225 epochs to 300 epochs. The mini-batch size $n=128$, momentum is 0.9 and weight decay is $5*10^{-4}$. From Figure~\ref{fig:epoch_alexnet}, we can find that in the classic AlexNet, the 3-$rd$ and 4-$nd$ convolutional layers' gaussianity have a decreasing trend with training. But batch normalization helps each convolutional layer to remain strong Gaussian behavior.

\begin{figure}[!]
\centering
\subfigure[Alexnet (200 batches)]{
\label{fig:200alexnet}
\includegraphics[width=0.475\columnwidth,height=3.1cm]{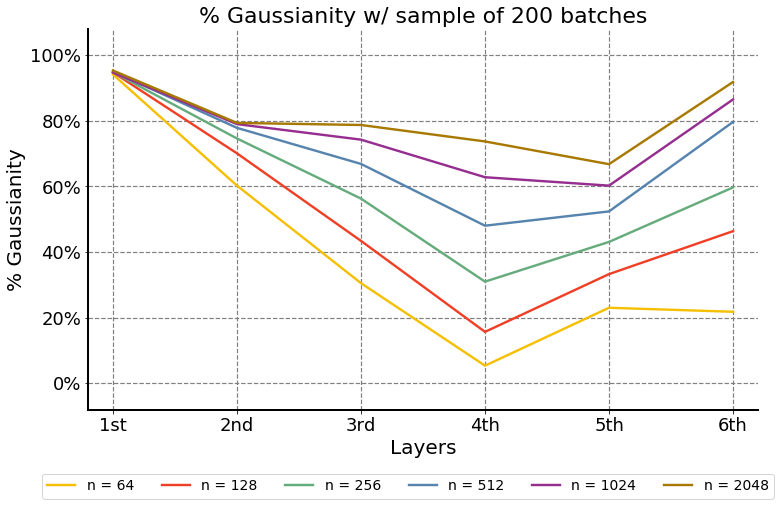}}
% \qquad
\subfigure[Alexnet (600 batches)]{
\label{fig:600alexnet}
\includegraphics[width=0.475\columnwidth,height=3.1cm]{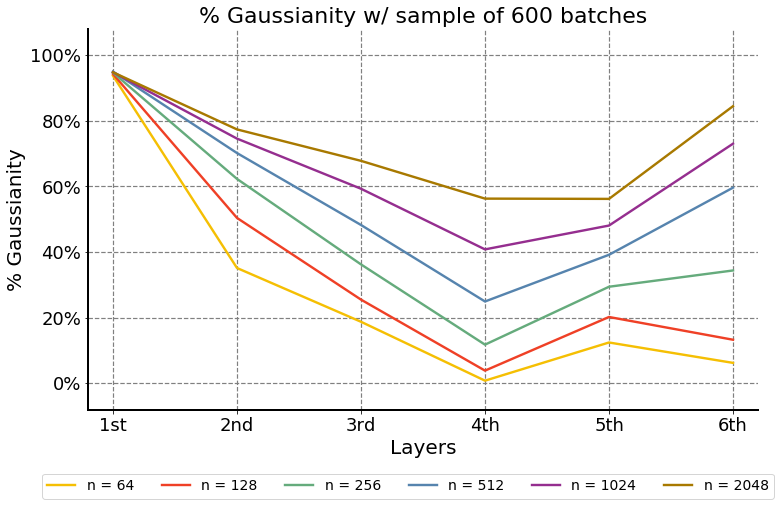}}
\vskip -5pt%
\caption{Gaussianity of each layer in Alexnet}
\vskip -10pt%
\end{figure}

To find more properties of gaussianity in the classic AlexNet, we did two experimental cases for two different batch numbers: 200 batches and 600 batches. In each case, we test several different batch size $n$ to see how it affects the Gaussianity of gradient noise. Figure~\ref{fig:200alexnet} and \ref{fig:600alexnet} show the same trend of Gaussianity of stochastic gradient noise. The percentage of Gaussianity has a steady increase with the larger batch size. The more details have been displayed in Appendix (Table~\ref{alexnet_table_200} and \ref{alexnet_table_600}). We show five convolutional layers' features weight sizes and the final layer's classifier weight size in the last column of the table. For each experiment, we have repeated five times to obtain the more accurate results, which can also be verified by the very small standard derivation values on the table.

Based on experimental results, we can find that there is a huge rise on the 4-$th$ convolutional layer, around zero percent Gaussianity with batch size $n=64$ goes up to $56.29\%$ (on 600 batches) and $73.73\%$ (on 200 batches) with batch size $n=2048$. They indicate the characteristics of gradient noise can be changed from non-Gaussian to Gaussian behavior if enough batch size is provided. The gradient noise on the first convolutional layer shows the best Gaussian behavior which covers all the batch size settings. When batch size $n=2048$, the gradient noise of the whole neural network method is above $50\%$ (on 600 batches) and $60\%$ (on 200 batches). These Gaussian behavior results satisfy our expectations and are consistent with the initial Gaussianity assumption.

From Figure~\ref{fig:200alexnet} we have known that the 1-$st$ layer's gaussianity percentage of AlexNet (mini-batch 128 and 200 batches) is 94.12\%. Shapiro-Wilk test still tells us that about five percent gradient noise does not have Gaussian behavior. Now we utilize tail index estimation~\footnote{https://github.com/josemiotto/pylevy} and fit the rest non-Gaussian behavior gradient noise to Levy alpha-stable distribution. In Appendix, we show the tail index performance on Figure~\ref{fig:tail_index_total} and show the example of gradient noise on Figure~\ref{fig:tail_index_non}. Figure~\ref{fig:tail_index_first} show the example of gradient noise of the 4-$th$ layer in AlexNet compared with different mini-batch settings.

\subsection{Gaussian behavior of ResNet}
\label{Resnet}

\begin{figure}[t]
\centering
\subfigure[ResNet (200 batches)]{
\label{fig:200resnet}
\includegraphics[width=0.475\columnwidth,height=3.1cm]{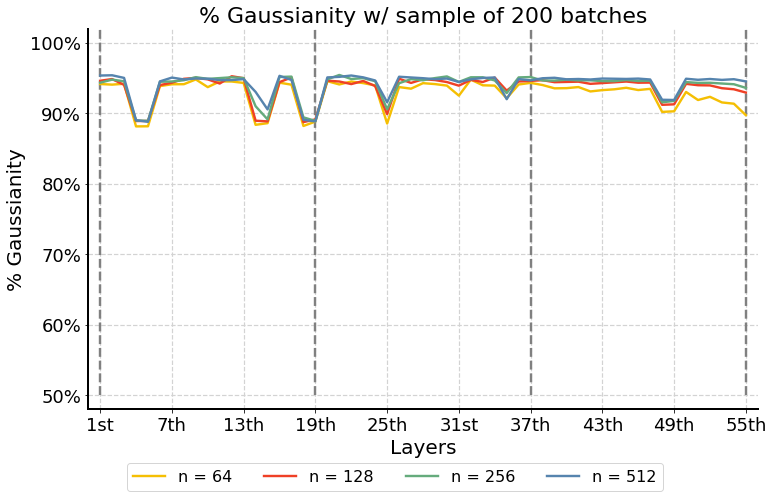}}
% \qquad
\subfigure[ResNet (600 batches)]{
\label{fig:600resnet}
\includegraphics[width=0.475\columnwidth,height=3.1cm]{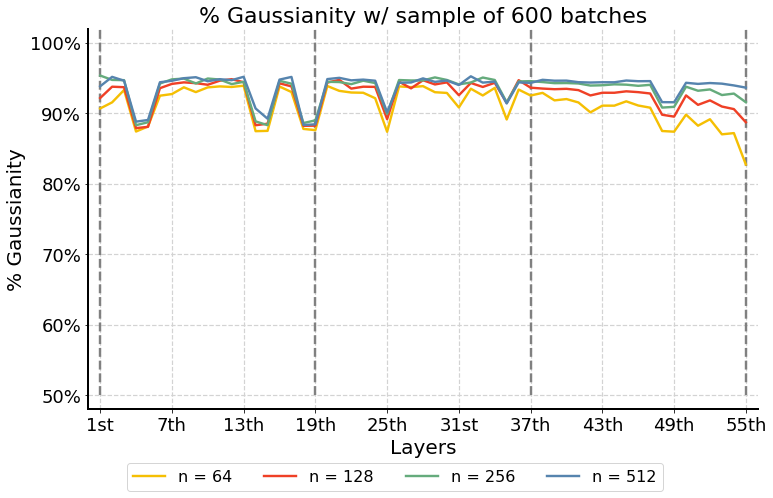}}
\vskip -5pt%
\caption{Gaussianity of each layer in ResNet-56}
\vskip -10pt%
\end{figure}

A Residual Neural Network (ResNet)~\cite{he2016deep} is also a widely-used artificial neural network, its structure utilizes skip connections, or shortcuts to jump over some layers. For implementation, we choose official ResNet-20 and ResNet-56 packages from Pytorch~\footnote{https://github.com/pytorch/vision}. The structure of ResNet is much more complicated, which contains the first convolutional layer, three block layers, and the final linear layer. Each block layer contains 6 convolutional layers with batch normalization for ResNet-20 or 18 convolutional layers with batch normalization for ResNet-56. 

We train ResNet-20 on Cifar10 datasets with 300 epochs; the gaussianity performance of each block layer's first convolutional layer and sixth convolutional layer during whole training epochs is shown in Appendix (Figure~\ref{fig:epoch_resnet}). ResNet has a very steady gaussianity for gradient noise, it does not change with training epochs. Furthermore, we do two experimental cases with ResNet-56 for two different batch numbers: 200 batches and 600 batches. Figure~\ref{fig:200resnet} and \ref{fig:600resnet} show ResNet-56 has a very high percentage of Gaussianity for gradient noise in each layer, which satisfies our Gaussianity assumption very well. Even if the batch size is small as $64$, they still show strong Gaussian behavior with above 85 percent Gaussianity. Besides, we can find similar results from AlexNet, which is the percentage of Gaussianity becomes higher with the larger batch size. The four light gray dashed lines divide three block layers on the plots. We also show these three block layers in Table~\ref{resnet_table_200} and \ref{resnet_table_600} in Appendix. In each block, for layout reasons, we only show layer 1, layer 9, and layer 18 in the table. We show the first convolutional layers' features weight dimension sizes, block layers' convolutional layers' features weight dimension sizes, and the final layer's classifier weight dimension size in the last column of the table. We repeat the experiment five times, the results are quite accurate with very small standard derivation errors. 

\subsection{Analysis of Bound}

According to Figure~\ref{fig:200alexnet} and \ref{fig:600alexnet}, we find that AlexNet's 4-$th$ and 5-$th$ convolutional layer show less Gaussian behavior in stochastic gradient noise compared with other layers. The rate of convergence may be the cause of such a phenomenon. We have introduced the Berry-Esseen theorem which states that the convergence of CLT is bounded by the standardized third absolute moment $\beta/\sigma^3$ and the summation size (batch size) $|\mathscr{S}|$. We keep as constant the batch size $|\mathscr{S}|$, and evaluate numerically the values of standardized third absolute moments $\beta/\sigma^3$ to measure the bound of convergence of CLT. We show in Appendix Figure~\ref{fig:alexnet_distribution} that the empirical distribution of the standardized third absolute moments of all weight dimensions within each layer of AlexNet. Compared with Figure~\ref{fig:200alexnet} and \ref{fig:600alexnet}, it is noticeable that 4-$th$ and 5-$th$ convolutional layers have longer right tails and generally get larger values of the standardized third absolute moment, which means their bound is much larger than other layers, so that slow convergence to Gaussian occurs. 

Looking into more details in Appendix for 4-$th$ convolutional layer from Figure~\ref{fig:alexnet_distribution}, we also estimate the quantile values of the third absolute moments distribution corresponding to batch size (from $n=64$ to $n=2048$ in Table~\ref{fig:200alexnet}) in Appendix Figure~\ref{fig:linevalues4}. For each batch size in this test, we calculate a quantile value w.r.t. the percentage of Gaussian dimensions. The results are demonstrated as vertical spines in Appendix Figure~\ref{fig:linevalues4}, the area on the left hand side corresponds roughly to those dimensions that converge towards Gaussian; the spines are shifting right as the batch size grows larger, indicates more dimensions exhibits Gaussian behavior.

Figure~\ref{fig:600resnet} shows that the percentage of Gaussianity in ResNet-56 Block-3 drops from $94.44\%$ to $91.58 \%$ when batch size $n=256$. We select convolutional layer 1 in Block-3  ($94.44\%$), layer 2 ($94.29\%$), layer 11 ($90.82\%$), layer 12 ($90.93\%$), layer 15 ($93.40\%$) and layer 16 ($92.60\%$) to show the empirical distribution of the standardized third absolute moments of weights in Appendix Figure~\ref{fig:resnet_distribution}; similar results has been shown as AlexNet. The layer with a higher Gaussianity of gradient noise often has a smaller value of bound. The CLT converges faster to Gaussian if the shape of the distribution of the standardized third absolute moments of weights is close to zero. Based on the comparison of Figure~\ref{fig:alexnet_distribution} and Figure~\ref{fig:resnet_distribution} in Appendix, we also notice the overall shape of distribution from ResNet is close to zero, only covers very small range around 0.01 to 0.02. It implies that ResNet has a strong Gaussian behavior for stochastic gradient noise. The overall shape of distribution from AlexNet is more flexible, the range reaches 0.2 in some cases. This explains why AlexNet has significant non-Gaussian behavior in some cases when batch size is not large enough.

\section{Conclusion}

In this paper, we investigated the characteristics of gradient noise during the training of deep neural networks using stochastic optimization methods. We showed that the gradient noise of stochastic methods exhibits finite variance during training. %the training of a typical neural network. 
The noise is therefore asymptotically Gaussian, instead of the latest proposal of $\alpha$-stable framework~\cite{simsekli2019tail}. To validate our claim that the non-Gaussianity is caused by the non-convergence of CLT, we conducted an empirical analysis on the percentage of Gaussian elements for each trainable layer and established quantitatively a connection of the Gaussianity with the standardized third absolute moment exploited in Berry-Esseen theorem. Our experiments show that the convergence rate of SGD noise is tied with the architecture of neural networks. The experiments show that the rate of convergence has a deep connection with the architecture of the neural networks: for each layer of AlexNet, the distributions of the third absolute moments calculated for all weight dimensions generally have a much wider spectrum, leading towards much worse Gaussianity in certain layers. For ResNet, the layer-wise moments' distributions have a narrower spectrum and smaller values, which validates the observations that the Gaussianity percentages of layers in ResNet are generally beyond $85\%$. When increasing the batch sizes, the percentage of Gaussianity for both AlexNet and ResNet improves. Based on the evidence of Gaussian noise, we analyze the dynamics of SGD and unify the discretization scheme w.r.t. the general Langevin equations: the vanilla SGD corresponds to the high-friction limit of the continuous-time dynamics. We then provide a simple local approximation of the distribution of learning parameters when the dynamics enter the vicinity of minima.

\clearpage

% \section{Acknowledgments}

\bigskip
\bibliography{gradient-noise}

\clearpage
\appendix
\onecolumn
\section*{Supplementary material for \\``Revisiting the Characteristics of Stochastic Gradient Noise and Dynamics''}
\subsection*{Supplementary to section ``Analysis of Gaussian Behavior''}
This section serves as a refinement of the proof, where we've proved that during the training of neural networks, the gradient noise will not admit infinite variance and therefore the (classic) CLT holds for asymptotic analysis. First, we show the completed proof of Lemma 1 and Lemma 2.

{\textbf{Lemma 1} 

\begin{proof}
By definition of Lipschitz continuation, we know
\begin{equation}
    ||f(x)-f(y)|| \le c||x-y||.
\end{equation}
For the any entry $f(x)_i=f(x_i)$ in $f(x)$, we have
\begin{align}
    \mathbf{var}[f(x_i)] &= \mathbf{var}[f(x_i)-f(\mathbb{E}[x_i])]\nonumber\\
                         &\le \mathbb{E}[(f(x_i)-f(\mathbb{E}[x_i]))^2]\nonumber\\
                         &\le c^2\mathbb{E}[(x_i-\mathbb{E}[x_i])^2]\nonumber\\
                         &= c^2\mathbf{var}[x_i].
\end{align}
For the random vector $f(x)$, the variance $\mathbf{var}[f(x)]$ will also be finite.
\end{proof}

}

{\textbf{Lemma 2}
\begin{proof}
The partial derivative with respect to the $i$-th entry $a^{(K)}_i$ in $a^{(K)}$ is
\begin{align}
\frac{\partial \mathscr{L}_{CE}}{\partial a^{(K)}_i} = -\sum_{j=1}^M y_j\frac{\partial}{\partial a^{(K)}_i}\log\frac{\exp{a^{(K)}_j}}{\sum_{k}\exp{a^{(K)}_k}} 
= -\sum_{j=1}^M y_j \frac{\sum_{k}\exp{a^{(K)}_k}}{\exp{a^{(K)}_j}}\frac{\partial}{\partial a^{(K)}_i}\frac{\exp{a^{(K)}_j}}{\sum_{k}\exp{a^{(K)}_k}}.\label{eq:ce_gradient}
\end{align}
Since
\begin{align}
\frac{\partial}{\partial a^{(K)}_i}\frac{\exp{a^{(K)}_j}}{\sum_{k}\exp{a^{(K)}_k}} &= \frac{\frac{\partial}{\partial a^{(K)}_i}\exp{a^{(K)}_j}\sum_{k}\exp{a^{(K)}_k}}{\left(\sum_{k}\exp{a^{(K)}_k}\right)^2}
\hspace{1em}-\frac{\exp{a^{(K)}_j}\frac{\partial}{\partial a^{(K)}_i}\sum_{k}\exp{a^{(K)}_k}}{\left(\sum_{k}\exp{a^{(K)}_k}\right)^2}\\
&= \begin{cases}
\frac{\exp{a^{(K)}_i}\sum_{k\neq i}\exp{a^{(K)}_k}}{\left(\sum_{k}\exp{a^{(K)}_k}\right)^2}, \quad (i=j)\\
-\frac{\exp{a^{(K)}_j}\exp{a^{(K)}_i}}{\left(\sum_{k}\exp{a^{(K)}_k}\right)^2}, \quad (i\neq j)
\end{cases}
\end{align}
we have
\begin{equation}
\begin{split}
\frac{\sum_{k}\exp{a^{(K)}_k}}{\exp{a^{(K)}_j}}\frac{\partial}{\partial a^{(K)}_i}\frac{\exp{a^{(K)}_j}}{\sum_{k}\exp{a^{(K)}_k}} =
\begin{cases}
\frac{\sum_{k\neq i}\exp{a^{(K)}_k}}{\sum_k\exp{a^{(K)}_k}}, (i=j)\\
-\frac{\exp{a^{(K)}_i}}{\sum_{k}\exp{a^{(K)}_k}}. (i\neq j)
\end{cases}\label{eq:exp_part}
\end{split} 
\end{equation}
Substituting Equation \eqref{eq:exp_part} into Equation \eqref{eq:ce_gradient}, we have
\begin{equation}
\left|\frac{\partial \mathscr{L}_{CE}}{\partial a^{(K)}_i}\right|\le y_i \frac{\sum_{k\neq i}\exp{a^{(K)}_k}}{\sum_k\exp{a^{(K)}_k}} + \sum_{j\neq i} y_j \frac{\exp{a^{(K)}_i}}{\sum_{k}\exp{a^{(K)}_k}}\le 1.
\end{equation}
In other words, the derivative w.r.t. $a^{(K)}$ is absolutely bounded by 1. 
As a result, $\partial \mathscr{L}_{CE}/\partial a^{(K)}$ has finite variance as the function of the network output $a^{(K)}$ and label $y$.
\end{proof}
}

Now we show why we need the assumption of finite fourth moments of the data distribution.
With the decomposition of variance, we have
\begin{equation}
\label{eq:proof}
\begin{split}
    \mathbf{var}\bigg[\frac{\partial \mathscr{L}}{\partial u^{(k)}}\frac{\partial u^{(k)}}{\partial W^{(k)}_{ij}}\bigg] 
    &= \mathbf{var}\bigg[\bigg[\frac{\partial \mathscr{L}}{\partial u^{(k)}}\bigg]_i a^{(k-1)}_j\bigg]\\
    &= \mathbb{E}\bigg[\bigg(\bigg[\frac{\partial \mathscr{L}}{\partial u^{(k)}}\bigg]_i a^{(k-1)}_j\bigg)^2\bigg] - \bigg(\mathbb{E}\bigg[\bigg[\frac{\partial \mathscr{L}}{\partial u^{(k)}}\bigg]_i a^{(k-1)}_j\bigg]\bigg)^2.
\end{split}
\end{equation}
With Cauchy inequality, we have
\begin{equation}
    \bigg(\mathbb{E}\bigg[\bigg[\frac{\partial \mathscr{L}}{\partial u^{(k)}}\bigg]_i a^{(k-1)}_j\bigg]\bigg)^2 \le \mathbb{E}\bigg[\bigg[\frac{\partial \mathscr{L}}{\partial u^{(k)}}\bigg]_i^2\bigg]\mathbb{E}\bigg[\bigg(a^{(k-1)}_j\bigg)^2\bigg].
\end{equation}
Recall $[\partial \mathscr{L}/\partial u^{(k)}]_i$ and $a^{(k-1)}_j$ have finite variance, which means the second term of RHS in Equation~\eqref{eq:proof} is also finite.
For the first term of RHS in Equation~\eqref{eq:proof}, we can evoke Cauchy inequality again and have
\begin{equation}
    \mathbb{E}\bigg[\bigg(\bigg[\frac{\partial \mathscr{L}}{\partial u^{(k)}}\bigg]_i a^{(k-1)}_j\bigg)^2\bigg] \le \sqrt{\mathbb{E}\bigg[\bigg[\frac{\partial \mathscr{L}}{\partial u^{(k)}}\bigg]_i^4\bigg]\mathbb{E}\bigg[\bigg(a^{(k-1)}_j\bigg)^4\bigg]}.
\end{equation}
With the same induction methodology adopted in section ``Analysis of Gaussian Behavior'', one can show that the 4-th moments of $[\partial \mathscr{L}/\partial u^{(k)}]_i$ and $a^{(k-1)}_j$ will also be finite by assuming the 4-th moments of data are finite.
Similarly, we know the 4-th moments of gradient noise $\mathbb{E}[\chi^4]$ is also finite under this assumption.
Moreover, by Jensen's inequality
\begin{equation}
    (\mathbb{E}[|\chi|^3])^{4/3} \le \mathbb{E}[(|\chi|^3)^{4/3}] = \mathbb{E}[\chi^4],
\end{equation}
which validates the usage of the Berry-Essen Theorem.\\

\subsection*{Supplementary to section ``Dynamics of SGD''}

Recall Equation~\eqref{eq:discret_langevin}, we denote $v_n = \Delta\theta_n$, $v_n - v_{n-1} = \Delta^2\theta_{n}$, the more details are shown as follow,

\begin{align}
 \gamma\frac{\Delta\theta_{n}}{\Delta{t}} &= -\nabla\mathscr{L}(\theta_{n-1}) - m\frac{\Delta^2\theta_{n}}{\Delta{t}^2}\\
 \Delta\theta_{n} &= - \frac{\Delta{t}\nabla\mathscr{L}(\theta_{n-1})}{\gamma} - \frac{m \Delta^2\theta_{n}}{\gamma \Delta{t}}\\
v_n &= - \frac{\Delta{t}\nabla\mathscr{L}(\theta_{n-1})}{\gamma} - \frac{m (v_n - v_{n-1})}{\gamma \Delta{t}}\\
v_n &= \frac{m}{\gamma \Delta{t} + m} v_{n-1} -  \frac{\Delta{t}^2}{\gamma \Delta{t} + m}\nabla\mathscr{L}(\theta_{n-1})\\
v_n &= \underset{\rho}{\underbrace{\frac{1}{1 + \gamma\Delta{t} / m}}} v_{n-1} - \underset{\alpha}{\underbrace{\frac{\Delta{t}}{\gamma + m/\Delta{t}}}} \nabla\mathscr{L}(\theta_{n-1}).
\end{align}
Recall the \emph{Langevin equation} \eqref{eq:langevin}, which we present as follows:
\begin{align}
\label{eq:a:langevin}
m\frac[1pt]{\diff^2{\theta}}{\diff{t}^2} + \gamma\frac[1pt]{\diff{\theta}}{\diff{t}} = -\nabla\phi(\theta) + \eta(\theta, t),
\end{align}
where $\theta \in \mathbb{R}^d$ denotes the positional state in a $d$-dimensional space, $\phi(\theta):\mathbb{R}^d \to \mathbb{R}$ is the potential function, and $\eta(t;\theta)$ the $\theta$-dependent $\delta$-correlated random noise with zero mean for every $\theta$ that
\begin{align}
\label{eq:a:eta}
\langle\,\eta(\theta, t)\,\rangle &\coloneq \mathbb{E}\big[ \,\eta(\theta, t) \,\big|\, \theta\, \big] \equiv 0, \notag\\
\langle\,\eta(\theta, t), \eta(\theta, t')\,\rangle &\coloneq \mathbb{E}\big[ \,\eta(\theta, t)\eta^\top(\theta, t') \,\big|\, \theta\, \big] = \mathsf{D}(\theta)\delta(t - t').
\end{align}
We cast the second-order \emph{underdamped Langevin equation} in \eqref{eq:a:langevin} to the corresponding stochastic differential equation (SDE) of first-order in vector form as
\begin{align}
\label{eq:a:sde}
\begin{bmatrix}[r]
\diff{\theta}\\
\diff{\varv}
\end{bmatrix} = \frac[1pt]{1}{m}
\begin{bmatrix}
\varv\\
-\gamma\varv - \nabla\phi(\theta)
\end{bmatrix} \diff{t} + \frac[1pt]{1}{m}
\begin{bmatrix}
0\\
\sqrt{\mathsf{D}(\theta)}
\end{bmatrix} \diff\mathcal{W},~\mbox{with the velocity}~\varv \coloneq \frac[1pt]{\diff{\theta}}{\diff{t}}~\mbox{and the standard Wiener process}~\diff{\mathcal{W}}.
\end{align}
By setting the mass as unity that $m = 1$ and defining the drift terms w.r.t. $\theta$ and $\varv$ as
\begin{align}
\label{eq:a:drift}
\mathcal{A}_{\theta} &= \varv~~\mbox{and}~~\mathcal{A}_{\varv} = -\big( \gamma\varv + \nabla\phi(\theta) \big),
\end{align}
the corresponding \emph{Fokker-Planck equation} (FPE) depicting the time evolution of the distribution $\pi(\theta, \varv, t)$ of states $(\theta, \varv)$ in the phase space reads
\begin{align}
\label{eq:a:fpe}
\frac[1pt]{\diff{\pi}(\theta, \varv, t)}{\diff{t}} = -\Big[ \nabla_{\theta} \cdot \big( \mathcal{A}_{\theta} \pi(\theta, \varv, t) \big) + \nabla_{\varv} \cdot \big( \mathcal{A}_{\varv} \pi(\theta, \varv, t) \big) \Big] + \frac[1pt]{1}{2} \Big[ \big( \nabla_{\varv}^\top \nabla_{\varv} \big) \odot \big( \mathsf{D}(\theta) \pi(\theta, \varv, t) \big) \Big],
\end{align}
where $\odot$ denotes the Hadamard (element-wise) product.

\noindent Note that the diffusion matrix $\mathsf{D}(\theta)$ is independent with the velocity $\varv$, we have a convenient formulation in the form of \emph{continuity equation} as (cf. \eqref{eq:j})
\begin{align}
\label{eq:a:j}
\frac[1pt]{\diff{\pi}}{\diff{t}} - \mathbf{div}\,\mathcal{J} = 0,~~\mbox{with the probability current}~~ \mathcal{J} \coloneq 
\begin{bmatrix}
\mathcal{J}_{\theta}\\
\mathcal{J}_{\varv}
\end{bmatrix} = -
\bigg\{
\begin{bmatrix}
\mathcal{A}_{\theta}\\
\mathcal{A}_{\varv}
\end{bmatrix} - \frac[1pt]{1}{2}
\begin{bmatrix}
0\\
\mathsf{D}(\theta) \cdot \nabla_{\varv}
\end{bmatrix}
\bigg\}\pi(\theta, \varv, t).
\end{align}
Now we recall the \emph{principle of detailed balance}~\cite{gardiner2009stochastic} that for when a system is in a stationary state, we have
\begin{align}
\label{eq:a:detailed_balance}
\pi^{\mathrm{ss}}(\theta, \varv) \mathbb{T}\big[(\theta, \varv) \to (\theta', \varv')\big] = \pi^{\mathrm{ss}}(\theta', -\varv') \mathbb{T}\big[(\theta', -\varv') \to (\theta, -\varv)\big],
\end{align}
where $\pi^{\mathrm{ss}}$ denotes the steady-state distribution while $\mathbb{T}$ represents the transition probability. A transition corresponds to a particle at some time $t$ with position velocity $(\theta, \varv)$ having acquired by a later time $t+\Delta t$ position and velocity $(\theta', \varv')$. Equation \eqref{eq:a:detailed_balance} essentially describes the balance between the transition $(\theta, \varv; t) \to (\theta', \varv'; t+\Delta t)$ and its time-reversed counterpart $(\theta', -\varv'; t+\Delta t) \to (\theta, -\varv; t)$ under a steady-state distribution of the system, which occurs \emph{if and only if} the probability current $\mathcal{J}$ in \eqref{eq:a:j} to vanish for all states $(\theta, \varv)$.
\\

\subsubsection{Proof of Proposition 1} 
\begin{proof}
By contradiction. Firstly, it can be read off directly from \eqref{eq:a:j} that in a stationary state, $\mathcal{J}_{\theta} = -\varv\,\pi^{\mathrm{ss}}(\theta, \varv) \neq 0$ for $\varv \neq 0$. Thus, the Kramers’ equation is not compatible with detailed balance~\cite{zhang2016hidden}. Secondly, suppose we have a steady-state solution to \eqref{eq:a:j} in the form of Gibbs distribution that
\begin{align}
\label{eq:a:gibbs}
\pi^{\mathrm{ss}}(\theta, \varv) \propto \exp\big[ -\psi(\theta, \varv) \big],
\end{align}
for some energy function $\psi$ of system states $(\theta, \varv)$. It implies \emph{potential condition} that
\begin{align}
\label{eq:a:potential_condition}
\nabla\nabla^\top{\psi} = \Big(\nabla\nabla^\top{\psi}\Big)^\top.
\end{align}
Here we propose the split of the drift $\mathcal{A}$ as
\begin{align}
\label{eq:a:split}
\mathcal{A} = \mathcal{A}^{(1)} + \mathcal{A}^{(2)},~~\mbox{with}~~ \mathcal{A}^{(1)} = 
\begin{bmatrix}
\mathcal{A}^{(1)}_{\theta}\\
\mathcal{A}^{(1)}_{\varv}
\end{bmatrix} = 
\begin{bmatrix}
0\\[2\jot]
-\gamma\varv
\end{bmatrix}~~\mbox{and}~~
\mathcal{A}^{(2)} = 
\begin{bmatrix}
\mathcal{A}^{(2)}_{\theta}\\
\mathcal{A}^{(2)}_{\varv}
\end{bmatrix} = 
\begin{bmatrix}
\varv\\[2\jot]
-\nabla\phi(\theta)
\end{bmatrix}.
\end{align}
Given $\mathsf{D}(\theta)$ being the covariance matrix from \eqref{eq:a:eta}, we presume it is positive definite, and hence invertible. Then we propose the ansatz such that
\begin{align}
\label{eq:a:psi}
\nabla_{\varv}{\psi}(\theta, \varv) = 2\mathsf{D}^{-1}(\theta) \big[ -\mathcal{A}^{(1)}_{\varv} \big],~~\mbox{thus}~~ \psi(\theta, \varv) = \gamma \int \mathsf{D}^{-1}(\theta)\,\varv\diff{\varv} = \gamma\varv^\top\mathsf{D}^{-1}(\theta)\,\varv + h(\theta).
\end{align}
As $\mathsf{D}^{-1}(\theta)$ being symmetric, it is straightforward to verify that $\psi(\theta, \varv)$ in the form of \eqref{eq:a:psi} satisfies the potential condition in \eqref{eq:a:potential_condition} in the $\varv$-space, we have
\begin{align}
\label{eq:a:div}
\mathbf{div} \left(\mathcal{A}^{(2)} e^{-\psi}\right) =
\mathbf{div} \left( 
\begin{bmatrix}
v\\
-\nabla\phi(\theta)
\end{bmatrix}
e^{-\psi} \right) &= \nabla_{\theta} \cdot \big[\varv e^{-\psi}\big] - \nabla_{\varv} \cdot \big[\nabla\phi(\theta)e^{-\psi}\big] \notag\\&= -\varv e^{-\psi} \cdot \left( \nabla_{\theta} \psi - 2\gamma\mathsf{D}^{-1}(\theta) \nabla\phi(\theta) \right) = 0.
\end{align}
Let \eqref{eq:a:div} to vanish, it has to be guaranteed that
\begin{align}
\label{eq:a:gradient}
\nabla_{\theta} \psi - 2\gamma\mathsf{D}^{-1}(\theta) \nabla\phi(\theta) = 0.
\end{align}
However, \eqref{eq:a:gradient} holds only when $\mathsf{D}(\theta) \equiv D\mathbf{I}$, as one may apply the potential condition \eqref{eq:a:potential_condition} to verify.
\end{proof}

\ \\

\noindent In discretizing the continuous-time stochastic differential equation \eqref{eq:a:langevin}, given small time step $\tau \ll 1$, we have
\begin{align}
\eta(\theta, t) \to \epsilon_{n}(\theta),\quad\mbox{with}\quad n = t/\tau.
\end{align}
The characteristics of the discrete-time noise $\epsilon_n(\theta)$ reads
\begin{align}
\mathbb{E}\big[\epsilon_n(\theta) \big| \theta\big] &\equiv 0,\notag\\ \mathbf{var}\big[\epsilon_n(\theta) \big| \theta\big] &= \mathsf{C}(\theta) \equiv \mathsf{D}(\theta)/\tau,
\end{align}
which implies that for small times $\tau$, we have $\mathsf{D}_{ij}(\theta) = \tau\mathsf{C}_{ij}(\theta) \ll 1$.

\subsubsection{Proof of Proposition 2} 
\begin{proof}
\noindent Given Assumption \ref{asm:c2}, the gradient $\nabla\phi(\theta)$ and hessian $\mathsf{H}(\theta)$ of $\phi(\theta)$ exist and are continuous. For $\theta$ within the vicinity of a local minimum $\theta_*$, the gradient $\nabla\phi(\theta) \approx 0$. Recall the definition of $\mathsf{C}(\theta)$ in \eqref{eq:hessian}, which we present as follows
\begin{align}
\label{eq:a:hessian}
\mathsf{C}(\theta) = \mathbf{cov}[\nabla\mathscr{L}(\theta)] = \nabla^2\phi(\theta) - \nabla\phi(\theta)\nabla^\top\phi(\theta) \approx \nabla^2\phi(\theta_*) = \mathsf{H}(\theta_*) = \mathsf{H}_*.
\end{align}
We apply the Taylor expansion to the second order as
\begin{align}
\label{eq:a:taylor}
\phi(\theta) \approx \phi(\theta_*) + \nabla^\top\phi(\theta_*) (\theta - \theta_*) + \frac[1pt]{1}{2} (\theta - \theta_*)^\top \mathsf{H}_* (\theta - \theta_*) = \phi(\theta_*) + \frac[1pt]{1}{2} (\theta - \theta_*)^\top \mathsf{H}_* (\theta - \theta_*).
\end{align}
The same split of the drift $\mathcal{A}$ as shown in \eqref{eq:a:split} and integrate the effective potential $\psi(\theta, \varv)$ such that
\begin{align}
\psi(\theta, \varv) = 2\gamma \int \mathsf{D}^{-1}(\theta)\,\varv\diff{\varv} = \frac[1pt]{\gamma}{\tau}\varv^\top\mathsf{C}^{-1}(\theta)\,\varv + h(\theta) \approx \frac[1pt]{\gamma}{\tau}\varv^\top\mathsf{H}_*^{-1}\,\varv + h(\theta).
\end{align}
We then proceed to solve for $h(\theta)$ by inserting \eqref{eq:a:div} and obtain
\begin{align}
\nabla_{\theta} \psi \approx \nabla h(\theta) = \frac[1pt]{2\gamma}{\tau}\mathsf{H}_*^{-1} \nabla\phi(\theta) \approx \frac[1pt]{2\gamma}{\tau}\mathsf{H}_*^{-1} \mathsf{H}_* (\theta - \theta_*) = \frac[1pt]{2\gamma}{\tau} (\theta - \theta_*).
\end{align}
By integration, we have
\begin{align}
h(\theta) = \frac[1pt]{2\gamma}{\tau} \int (\theta - \theta_*) \diff{\theta} = \frac{\gamma}{\tau} (\theta - \theta_*)^\top (\theta - \theta_*).
\end{align}
Substitute $h(\theta)$ back into \eqref{eq:a:gibbs} we have
\begin{align}
\label{eq:a:steady-state}
\pi^{\mathrm{ss}} \propto \exp\Big[ -\frac[1pt]{\gamma}{\tau} (\theta - \theta_*)^\top(\theta - \theta_*) \Big] \exp\Big[ -\frac[1pt]{\gamma}{\tau}\varv^\top \mathsf{H}_*^{-1}\varv \Big].
\end{align}
\end{proof}

\clearpage
\subsection*{Supplemental tables}

\begin{table*}[h]
\centering
\caption{Gaussianity percentage of each layer in Alexnet (200 batches)}
\label{alexnet_table_200}
\begin{adjustbox}{width=16cm}
\begin{tabular}{|c|c|c|c|c|c|c|c|}
\hline
200 Batches & \multicolumn{6}{c|}{Gaussianity percentage (\%)}                                                    &             \\ \cline{1-7}
Batch Size  & 64             & 128            & 256            & 512            & 1024           & 2048           & Weight Size \\ \hline
Layer 1     & $94.13\pm0.36$ & $94.79\pm0.21$ & $94.85\pm0.18$ & $95.02\pm0.25$ & $94.75\pm0.52$ & $95.34\pm0.23$ & 23,232      \\ \hline
Layer 2     & $60.25\pm0.86$ & $70.09\pm0.83$ & $74.62\pm0.46$ & $77.86\pm1.13$ & $79.00\pm0.68$ & $79.37\pm0.30$ & 307,200     \\ \hline
Layer 3     & $30.52\pm0.21$ & $43.41\pm0.22$ & $56.30\pm0.26$ & $66.87\pm0.16$ & $74.27\pm0.14$ & $78.74\pm0.19$ & 663,552     \\ \hline
Layer 4     & $5.37\pm0.09$  & $15.68\pm0.09$ & $30.98\pm0.25$ & $48.04\pm0.32$ & $62.81\pm0.13$ & $73.72\pm0.12$ & 884,736     \\ \hline
Layer 5     & $23.02\pm0.24$ & $33.28\pm0.22$ & $43.09\pm0.14$ & $52.39\pm0.16$ & $60.25\pm0.11$ & $66.79\pm0.14$ & 589,824     \\ \hline
Layer 6     & $21.80\pm4.11$ & $46.37\pm1.91$ & $59.70\pm5.20$ & $79.66\pm5.78$ & $86.52\pm3.89$ & $91.84\pm1.18$ & 2,560       \\ \hline
\end{tabular}
\end{adjustbox}
\end{table*}

\begin{table*}[h]
\centering
\caption{Gaussianity percentage of each layer in Alexnet (600 batches)}
\label{alexnet_table_600}
\begin{adjustbox}{width=16cm}
\begin{tabular}{|c|c|c|c|c|c|c|c|}
\hline
600 Batches & \multicolumn{6}{c|}{Gaussianity percentage (\%)}                                                    & \multicolumn{1}{l|}{} \\ \cline{1-7}
Batch Size  & 64             & 128            & 256            & 512            & 1024           & 2048           & Weight Size           \\ \hline
Layer 1     & $93.74\pm0.53$ & $94.27\pm0.28$ & $94.70\pm0.11$ & $95.02\pm0.36$ & $94.82\pm0.32$ & $94.81\pm0.22$ & 23,232                \\ \hline
Layer 2     & $35.14\pm0.79$ & $50.41\pm1.07$ & $62.30\pm0.31$ & $70.18\pm0.16$ & $74.58\pm0.46$ & $77.39\pm0.36$ & 307,200               \\ \hline
Layer 3     & $18.75\pm0.14$ & $25.50\pm0.19$ & $36.24\pm0.23$ & $48.26\pm0.12$ & $59.30\pm0.20$ & $67.81\pm0.16$ & 663,552               \\ \hline
Layer 4     & $0.80\pm0.06$  & $3.86\pm0.08$  & $11.78\pm0.09$ & $24.93\pm0.20$ & $40.82\pm0.13$ & $56.29\pm0.12$ & 884,736               \\ \hline
Layer 5     & $12.44\pm0.18$ & $20.21\pm0.15$ & $29.43\pm0.14$ & $39.17\pm0.27$ & $48.08\pm0.16$ & $56.21\pm0.12$ & 589,824               \\ \hline
Layer 6     & $6.22\pm0.62$  & $13.27\pm2.83$ & $34.39\pm6.29$ & $59.63\pm4.33$ & $73.03\pm2.98$ & $84.47\pm2.51$ & 2,560                 \\ \hline
\end{tabular}
\end{adjustbox}
\end{table*}
\begin{table*}[!]
\centering
\caption{Gaussianity percentage of each layer in ResNet (200 batches)}
\label{resnet_table_200}
\begin{adjustbox}{width=12.5cm}
\begin{tabular}{|c|c|c|c|c|c|c|}
\hline
\multicolumn{2}{|c|}{200 Batches} & \multicolumn{4}{c|}{Gaussianity percentage (\%)}                  &             \\ \cline{1-6}
\multicolumn{2}{|c|}{Batch Size}  & 64             & 128            & 256            & 512            & Weight Size \\ \hline
\multicolumn{2}{|c|}{First Layer} & $94.17\pm1.39$ & $94.63\pm0.40$ & $94.35\pm2.60$ & $95.37\pm2.25$ & 432         \\ \hline
                & Layer 1         & $94.08\pm0.65$ & $94.87\pm0.91$ & $94.79\pm0.54$ & $95.41\pm0.61$ & 2,304       \\
Block 1         & Layer 9         & $93.72\pm0.98$ & $94.83\pm0.29$ & $94.90\pm0.55$ & $94.92\pm0.36$ & 2,304       \\
                & Layer 18        & $88.84\pm0.68$ & $89.15\pm0.45$ & $88.93\pm0.64$ & $88.85\pm0.45$ & 2,304       \\ \hline
                & Layer 1         & $94.57\pm0.44$ & $94.64\pm0.39$ & $94.83\pm0.25$ & $95.10\pm0.46$ & 4,608       \\
Block 2         & Layer 9         & $94.29\pm0.26$ & $94.84\pm0.59$ & $94.70\pm0.39$ & $94.99\pm0.56$ & 9,216       \\
                & Layer 18        & $94.33\pm0.47$ & $94.61\pm0.35$ & $95.16\pm0.82$ & $94.69\pm0.28$ & 9,216       \\ \hline
                & Layer 1         & $94.02\pm0.12$ & $94.71\pm0.14$ & $94.67\pm0.29$ & $94.98\pm0.18$ & 18,432      \\
Block 3         & Layer 9         & $93.31\pm0.31$ & $94.35\pm0.07$ & $94.60\pm0.11$ & $94.94\pm0.09$ & 36,864      \\
                & Layer 18        & $89.75\pm1.03$ & $92.96\pm0.39$ & $93.62\pm0.45$ & $94.54\pm0.50$ & 36,864      \\ \hline
\multicolumn{2}{|c|}{Final Layer} & $41.62\pm9.53$ & $70.53\pm5.68$ & $75.72\pm5.96$ & $89.09\pm3.32$ & 640         \\ \hline
\end{tabular}
\end{adjustbox}
\end{table*}

\begin{table*}[!]
\centering
\caption{Gaussianity percentage of each layer in ResNet (600 batches)}
\label{resnet_table_600}
\begin{adjustbox}{width=12.5cm}
\begin{tabular}{|c|c|c|c|c|c|c|}
\hline
\multicolumn{2}{|c|}{600 Batches} & \multicolumn{4}{c|}{Gaussianity percentage (\%)}                  & \multicolumn{1}{l|}{} \\ \cline{1-6}
\multicolumn{2}{|c|}{Batch Size}  & 64             & 128            & 256            & 512            & Weight Size           \\ \hline
\multicolumn{2}{|c|}{First Layer} & $90.96\pm2.69$ & $92.18\pm2.70$ & $95.37\pm1.67$ & $93.84\pm0.68$ & 432                   \\ \hline
                & Layer 1         & $91.56\pm0.49$ & $93.78\pm0.91$ & $94.75\pm0.50$ & $95.18\pm0.57$ & 2,304                 \\
Block 1         & Layer 9         & $93.69\pm0.85$ & $94.06\pm0.63$ & $94.95\pm0.70$ & $94.56\pm0.44$ & 2,304                 \\
                & Layer 18        & $87.62\pm0.41$ & $88.26\pm0.84$ & $89.01\pm0.63$ & $88.45\pm0.49$ & 2,304                 \\ \hline
                & Layer 1         & $93.88\pm0.52$ & $94.38\pm0.60$ & $94.48\pm0.18$ & $94.87\pm0.32$ & 4,608                 \\
Block 2         & Layer 9         & $93.86\pm0.30$ & $94.70\pm0.50$ & $94.68\pm0.47$ & $94.97\pm0.87$ & 9,216                 \\
                & Layer 18        & $92.53\pm0.46$ & $93.65\pm0.55$ & $94.57\pm0.50$ & $94.32\pm0.46$ & 9,216                 \\ \hline
                & Layer 1         & $92.91\pm0.30$ & $93.52\pm0.20$ & $94.44\pm0.17$ & $94.76\pm0.11$ & 18,432                \\
Block 3         & Layer 9         & $91.12\pm0.34$ & $93.01\pm0.30$ & $93.91\pm0.11$ & $94.56\pm0.25$ & 36,864                \\
                & Layer 18        & $82.75\pm0.94$ & $88.74\pm0.97$ & $91.58\pm0.70$ & $93.65\pm0.37$ & 36,864                \\ \hline
\multicolumn{2}{|c|}{Final Layer} & $9.50\pm2.15$  & $26.81\pm3.96$ & $59.44\pm3.85$ & $75.72\pm6.02$ & 640                   \\ \hline
\end{tabular}
\end{adjustbox}
\end{table*}

\clearpage
\subsection*{Supplemental figures}

\begin{figure*}[h]
\centering
\subfigure[AlexNet]{
\label{fig:alexnet_distribution}
\includegraphics[width=6.3cm,height=10.35cm]{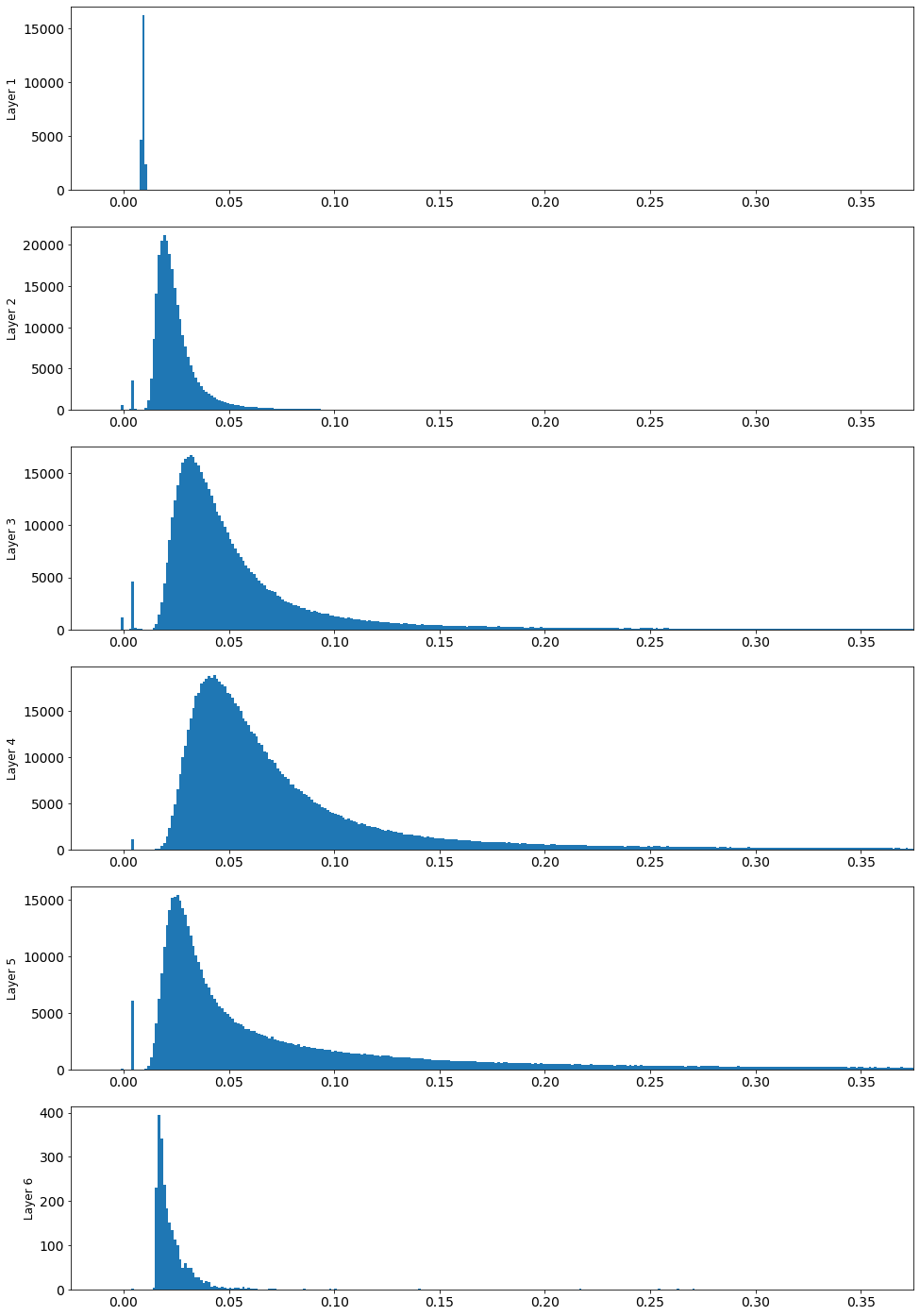}}
\subfigure[ResNet]{
\label{fig:resnet_distribution}
\includegraphics[width=6.3cm,height=10.35cm]{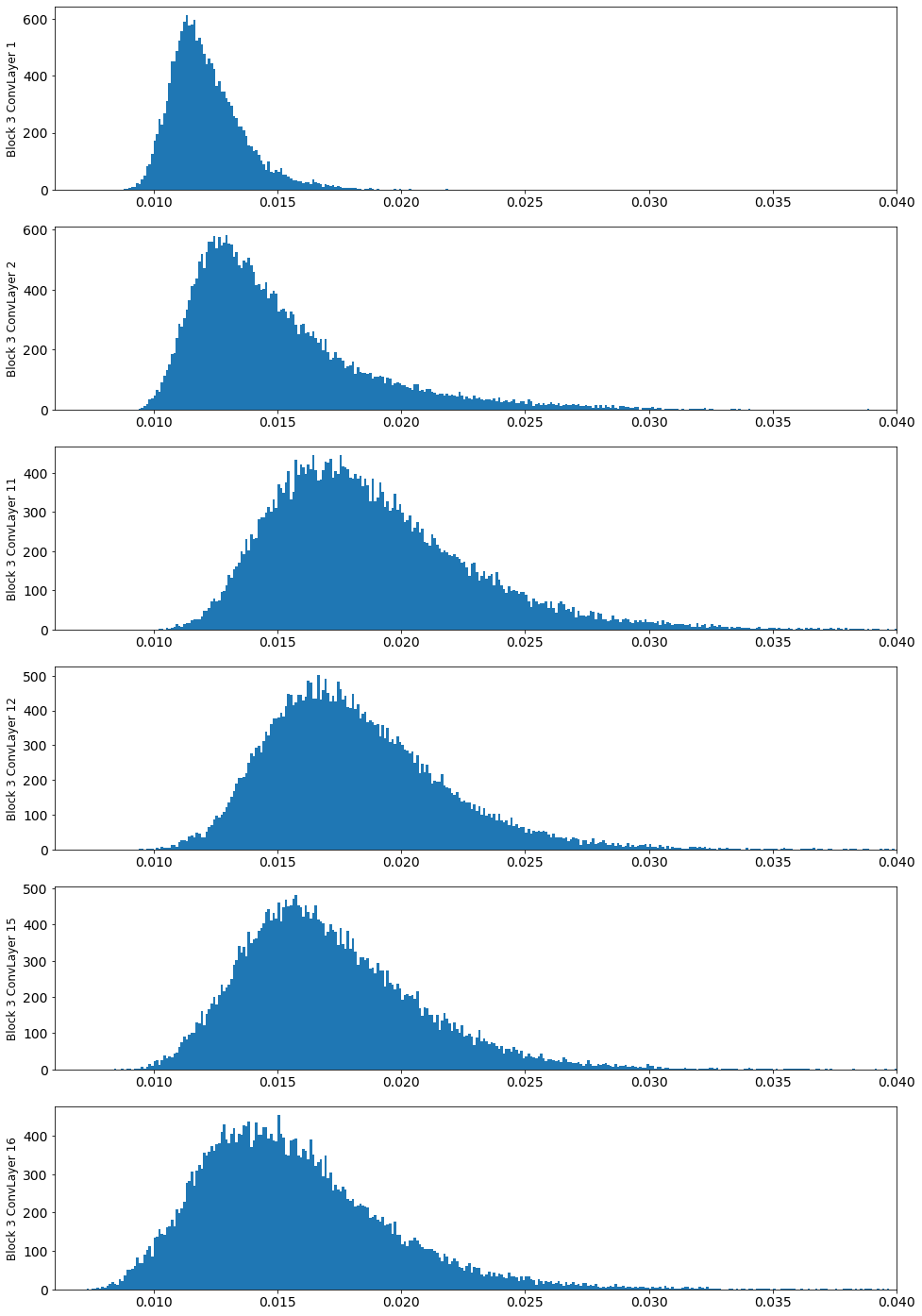}}
\caption{Empirical distribution of the standardized third absolute moments of weights on AlexNet and ResNet.}
\end{figure*}

Figure~\ref{fig:alexnet_distribution} shows the empirical distribution of the standardized third absolute moments of weights on AlexNet (cf. Figure~\ref{fig:200alexnet} and Figure~\ref{fig:600alexnet}). Figure~\ref{fig:resnet_distribution} shows the empirical distribution of the standardized third absolute moments of weights on ResNet (cf. Figure~\ref{fig:200resnet} and Figure~\ref{fig:600resnet}).

\begin{figure}[h]
\centering
\subfigure[3-$rd$ Layer]{
\label{fig:linevalues3}
\includegraphics[width=8.5cm]{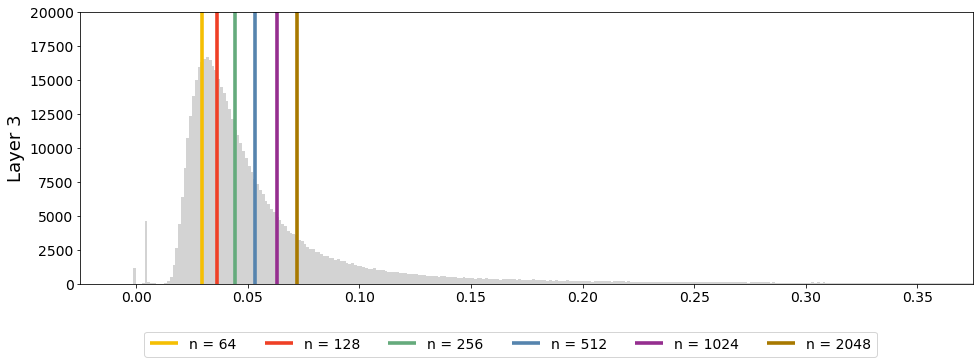}}
\qquad
\subfigure[4-$th$ Layer]{
\label{fig:linevalues4}
\includegraphics[width=8.5cm]{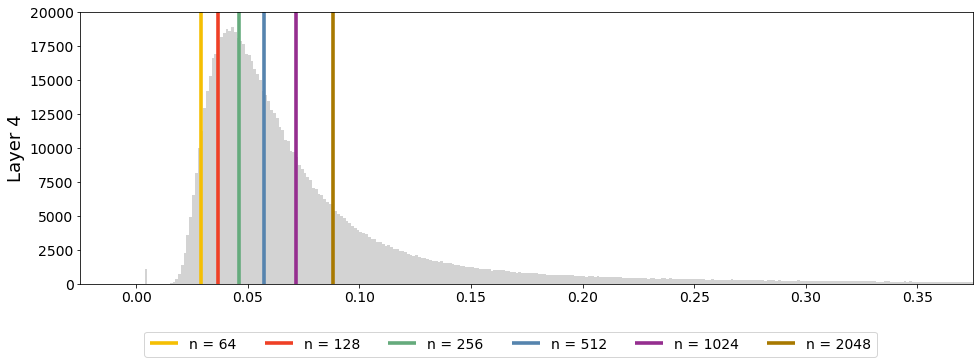}}
\qquad
\subfigure[5-$th$ Layer]{
\label{fig:linevalues5}
\includegraphics[width=8.5cm]{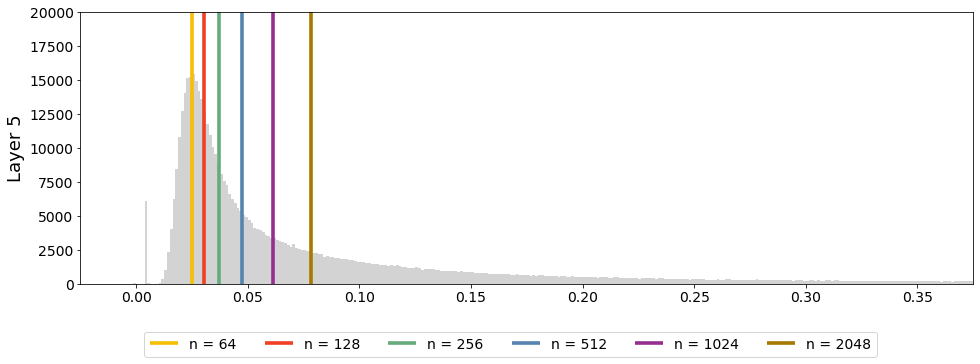}}
\caption{Quantile values on standardized third absolute moments of weights on AlexNet.}
\label{fig:linevalues}
\end{figure}

\newpage
Figure~\ref{fig:linevalues5} shows the distributions of the Berry-Esseen bound with quantile bars. The histograms are the empirical distributions of the element-wise bound for specific layers in AlexNet, where the greater values of the bound indicates a slower convergence rate towards the Gaussian according to the (classic) CLT.

\begin{figure}[!]
\centering
\includegraphics[width=8.5cm]{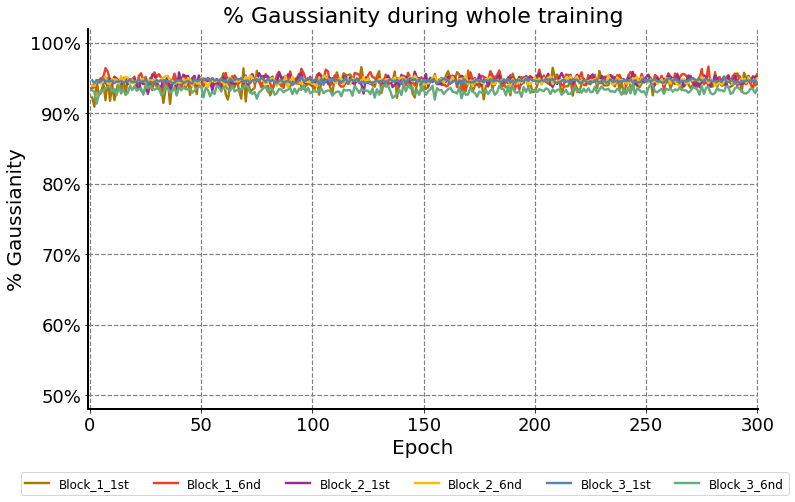}
\caption{Gaussianity of each layer in ResNet-20 during training epochs.}
\label{fig:epoch_resnet}
\end{figure}

\newpage
Recent research works on gradient noise normally employ some tail index measurement, predominately estimation of the $\alpha$ parameter in the $\alpha$-stable distribution, to support the argument of heavy tail behavior.
The parameter $\alpha$ (with range (0,2)) in $\alpha$-stable distributions is the parameter to determine the tail behavior of them, where the smaller the parameter, heavier the tail will be.
However, by using tail index measurement for $\alpha$-stable distributions, one \textit{a priori} assumes the distribution is in the $\alpha$-stable family, which has infinite long tail and infinite variance, and aims at estimating how heavy the infinite tail is with the empirical distribution of finite many data.
On the contrast, we start from the assumption on data, which is reasonable for realistic data, e.g. natural images, and prove that the variance of gradient noise will be not infinite.

\begin{figure}[th]
\centering
\includegraphics[width=8cm]{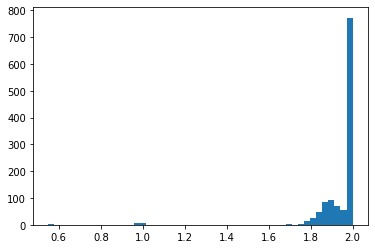}
\caption{Tail index performance of gradient noise with non-Gaussian behavior.}
\label{fig:tail_index_total}
\end{figure}

\begin{figure*}[th]
\centering
\subfigure[$n=128$;$\alpha=2.0$;
$\textrm{p-value}=0.724$]{
\includegraphics[width=2.63cm]{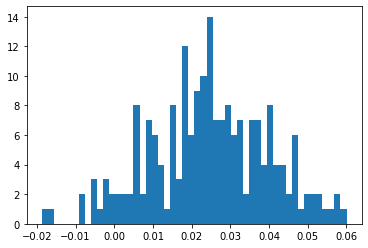}}
\subfigure[$n=128$;$\alpha=2.0$; $\textrm{p-value}=0.995$]{
\includegraphics[width=2.63cm]{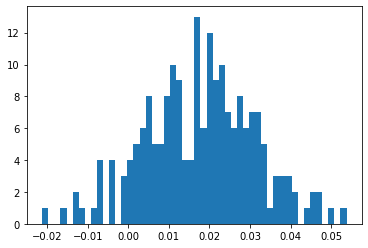}}
\subfigure[$n=128$;$\alpha=2.0$; $\textrm{p-value}=0.873$]{
\includegraphics[width=2.63cm]{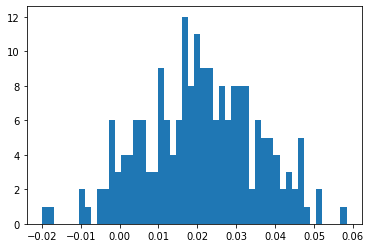}}
\subfigure[$n=128$;$\alpha=2.0$; $\textrm{p-value}=0.913$]{
\includegraphics[width=2.63cm]{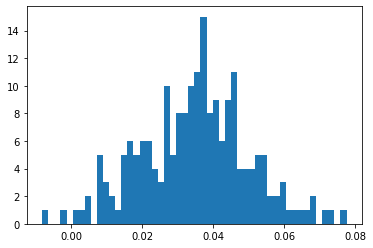}}
\subfigure[$n=128$;$\alpha=2.0$; $\textrm{p-value}=0.727$]{
\includegraphics[width=2.63cm]{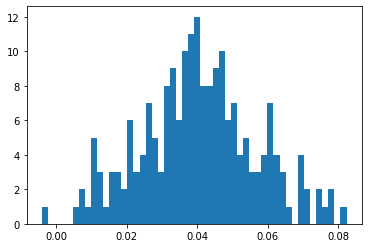}}

\subfigure[$n=128$;$\alpha=0.55$;
$\textrm{p-value}=8.16*10^{-3}$]{
\includegraphics[width=2.63cm]{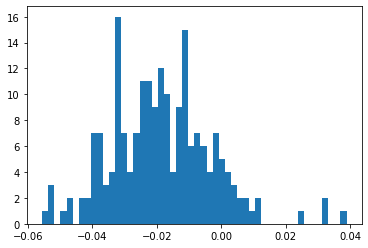}}
\subfigure[$n=128$;$\alpha=0.99$;
$\textrm{p-value}=2.78*10^{-3}$]{
\includegraphics[width=2.63cm]{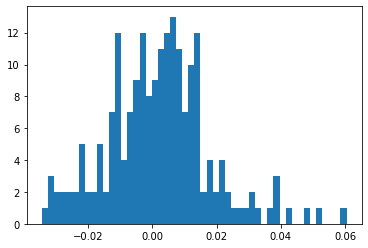}}
\subfigure[$n=128$;$\alpha=1.01$;
$\textrm{p-value}=0.032$]{
\includegraphics[width=2.63cm]{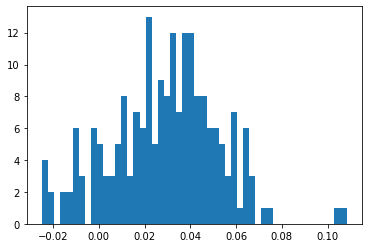}}
\subfigure[$n=128$;$\alpha=1.86$;
$\textrm{p-value}=0.021$]{
\includegraphics[width=2.63cm]{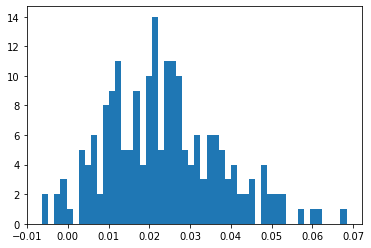}}
\subfigure[$n=128$;$\alpha=2.0$;
$\textrm{p-value}=0.010$]{
\includegraphics[width=2.63cm]{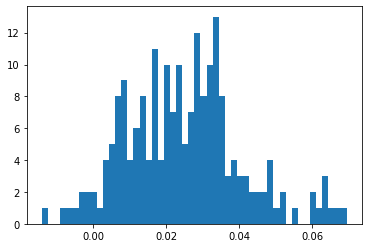}}
\caption{Example of gradient noise on the 1-$st$ layer of AlexNet under mini-batch $n=128$. From (a) to (e) we show the distribution of gradient noise with gaussian behavior by Shapiro-Wilk test (p-value is larger than 0.05), their tail index $\alpha$ all equal to 2 which satisfies Gaussian behavior. Based on Shapriro-Wilk test we also get small part of gradient noise which is not Gaussian, we test tail index properties of them and show their distribution from (f) to (j).}
\label{fig:tail_index_non}
\end{figure*}

\begin{figure*}[th]
\centering
\subfigure[$n=64$; $\alpha=0.5$; $\textrm{p-value}=1.71*10^{-8}$]{
\includegraphics[width=2.63cm]{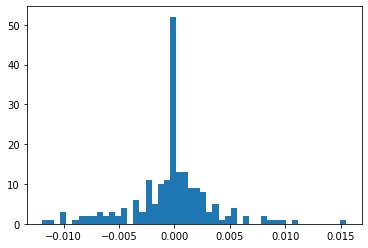}}
\subfigure[$n=128$;$\alpha=1.9$; $\textrm{p-value}=0.031$]{
\includegraphics[width=2.63cm]{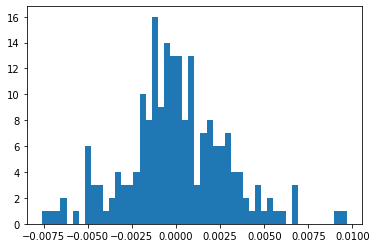}}
\subfigure[$n=256$;$\alpha=2.0$; $\textrm{p-value}=0.138$]{
\includegraphics[width=2.63cm]{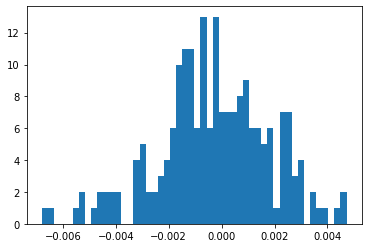}}
\subfigure[$n=512$;$\alpha=2.0$; $\textrm{p-value}=0.241$]{
\includegraphics[width=2.63cm]{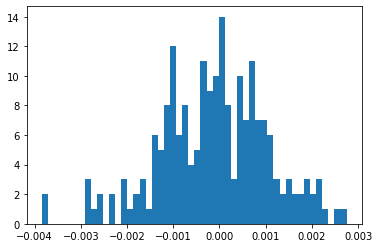}}
\subfigure[$n=1024$;$\alpha=2.0$; $\textrm{p-value}=0.969$]{
\includegraphics[width=2.63cm]{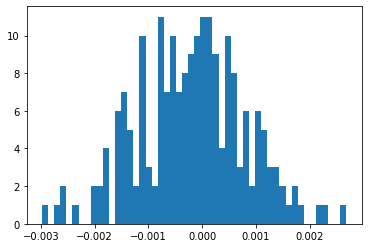}}
\caption{Example of gradient noise on the 4-$th$ layer of AlexNet with mini-batch $n$ from 64 to 1024.}
\label{fig:tail_index_first}
\end{figure*}

Nevertheless, we conducted related experiments and analyze the results with tail index measurements.
Figure \ref{fig:tail_index_total} shows the histogram of tail indices of gradient noise in one layer of AlexNet.
One can observe that most of the indices are 2 and most of the remaining ones are very close to 2.
Keep in mind that we do not use tail index to evaluate if the variance is infinite or not.
We use this experiment to show that, even if one thinks there are infinite long tails, the tails are still mostly as light as Gaussian.
Further, we show in Figure \ref{fig:tail_index_non} the histogram of gradient noise of different entries in the neural network parameter.
Comparing to those with $\alpha$ value 2.0, those with small $\alpha$ values tend to have groups of similar values on the tails.
These flat groups exhibit the heavy-tailness of these distributions.
However, since the values of gradient noise are bounded, the tails are also bounded, despite of the fact of slow decay in that part.
Moreover, the tails will have a sharp decay near the largest possible absolute value under certain dataset and at one certain training step.
As a result, one may not be able to conclude that the variances are infinite with evaluated $\alpha$ values smaller than 2.
From Figure \ref{fig:tail_index_first}, we can see that with larger batch size, the $\alpha$ value increase to 2.
Even if one employs $\alpha$-stable distribution for tail asymptotic analysis, Gaussian behavior will emerge with sufficient large batch size, which can also be taken as an evidence to support the noise is not intrinsically non-Gaussian.

\newpage

\end{document}